\documentclass[journal,twoside,web]{IEEEtran} 
\pdfoutput=1
\usepackage{cite}
\usepackage{amsmath,amssymb,amsfonts}
\usepackage{algorithmic}
\usepackage{graphicx}
\usepackage{textcomp}

\usepackage{url}
\usepackage{xcolor}
\usepackage{hyperref}
\usepackage{longtable,booktabs}
\usepackage{subcaption}
\usepackage[export]{adjustbox}
\usepackage{lineno}
\usepackage{bbding}  
\usepackage{array}
\usepackage[subtle,margins=normal,leading=normal]{savetrees} 
\usepackage{lettrine} 
\usepackage{soul}
\usepackage{cleveref}  

\usepackage[final]{changes}  

\begin{document}

\title{Do Gradient Inversion Attacks Make Federated Learning Unsafe?}
\author{Ali Hatamizadeh, Hongxu Yin, Pavlo Molchanov, Andriy Myronenko, Wenqi Li, Prerna Dogra, \\Andrew Feng, Mona G. Flores, Jan Kautz, Daguang Xu, and Holger R. Roth

\thanks{Submitted: \today}
\thanks{All authors are with NVIDIA, Santa Clara, USA. Contact Holger Roth (e-mail: \href{mailto:hroth@nvidia.com}{hroth@nvidia.com}). Code is available at \url{https://nvidia.github.io/NVFlare/research/quantifying-data-leakage}.}}

\maketitle

\begin{abstract}
Federated learning (FL) allows the collaborative training of AI models without needing to share raw data. This capability makes it especially interesting for healthcare applications where patient and data privacy is of utmost concern. However, recent works on the inversion of deep neural networks from model gradients raised concerns about the security of FL in preventing the leakage of training data. 
In this work, we show that these attacks presented in the literature are impractical in FL use-cases where the clients' training involves updating the Batch Normalization (BN) statistics and provide a new baseline attack that works for such scenarios. 
Furthermore, we present new ways to measure and visualize potential data leakage in FL.
Our work is a step towards establishing reproducible methods of measuring data leakage in FL and could help determine the optimal tradeoffs between privacy-preserving techniques, such as differential privacy, and model accuracy based on quantifiable metrics.
\end{abstract}



\section{Introduction}
\label{sec:introduction}

\lettrine[lines=2,findent=2pt]{\textbf{F}}{ederated} learning (FL) allows the collaborative training of Artificial Intelligence (AI) models without the need to share raw data~\cite{rieke2020future}. Participating client sites only share model weights or their updates with a centralized server and therefore, the underlying data privacy is being preserved~\cite{mcmahan2017communication}. This makes FL especially attractive for healthcare applications where patient privacy is of utmost concern. Several real-world applications of FL in medical imaging have shown its potential to improve the model performance and generalizability of AI models by allowing them to learn from large and diverse patient populations, which could include variations in scanner equipment and acquisition protocols, often spanning different countries, continents, and patient populations~\cite{sheller2018multi,roth2020federated,sheller2020federated,remedios2020federated,sarma2021federated,dayan2021federated}. FL has attracted a lot of attention from researchers~\cite{kairouz2019advances}, who study how to deal effectively with heterogeneous (or non-i.i.d.) data~\cite{kairouz2019advances,li2020federated,karimireddy2020scaffold}, address fairness and bias concerns~\cite{zhou2021towards}, improve the efficiency of client-server communication~\cite{rothchild2020fetchsgd}, and more~\cite{li2020federated}.

\begin{figure*}[htbp]
    \centering
    \includegraphics[width=1.5\columnwidth]{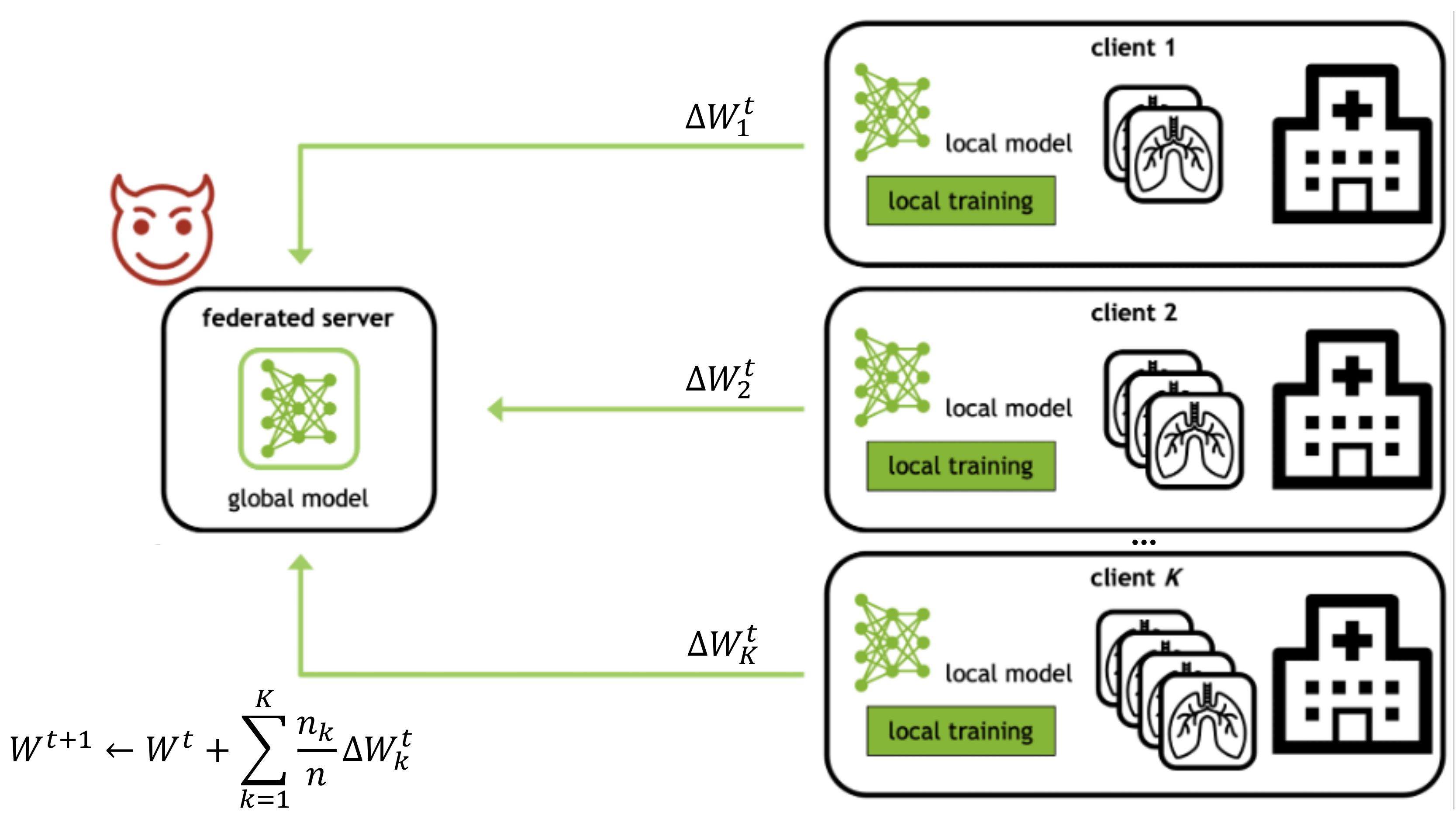}
    \caption{\textbf{The server-side gradient inversion attack scenario in FL.} In a typical FL setup for healthcare applications, $K$ participating sites (often hospitals or research institutions) keep their data locally while training a common global model. This model is distributed by a centralized federated server to initialize the current round $t$ of local training at each site. After the local training is completed, all sites or a subset of the sites send model updates $\Delta W_k^t$ with respect to the global model $W^t$ to the server, which aggregates them by performing a weighted average. The server then uses the aggregate to update the global model for the next round of FL ($t+1$). This iterative procedure continues until convergence of the models. Here, $n_k$ and $n$ stand for the number of images at client $k$ and the total number of images across clients, respectively. To study the gradient inversion attack, we assume an ``honest but curious''~\cite{bonawitz2017practical} scenario where the server performs the FL computations faithfully but could be compromised by a potential malicious actor who could gain access to the training configurations and model updates $\Delta W_k^t$ from the participating sites and attempt an inversion attack given the current state of the global model $W^t$ (see Section \ref{sec:methods}). \label{fig:attack_scenario}}
\end{figure*}

But how true is the assumption that FL preserves privacy? Several efforts have demonstrated the possibility of recovering the underlying training data (e.g., high-resolution images and corresponding labels) by exploiting shared model gradients (i.e., the model updates sent to the server during FL)~\cite{geiping2020inverting,yin2021see,kaissis2021end}. This attack scenario, known as the ``gradient inversion'' attack, could happen on both the server- and client-side during FL training. In typical FL scenarios for healthcare, the server-side gradient inversion attack assumes that all the clients trust each other~\cite{rieke2020future} while the server is ``honest but curious''~\cite{bonawitz2017practical}. This means the server follows the FL procedure faithfully but might be compromised and attempts to reconstruct some raw data from the clients (see Fig. \ref{fig:attack_scenario}). 

In the ``cross-silo'' scenario of FL~\cite{kairouz2019advances}, where an AI model is trained on a relatively small number of client sites with large amounts of data, one can assume that each client is running a relatively large number of training iterations on their local data before sending the model updates to the server. This is in contrast to the ``cross-device'' setting with many hundreds or thousands of clients, each with relatively small amounts of data, local compute resources, and unreliable network connections which might be dropping in and out at any time during training. In healthcare, there are different ways the data could be organized across institutes. In the so-called ``horizontal'' FL, each institute has data from different patients who share common attributes for a certain modeling task. On the other hand, ``vertical'' FL allows institutes which share a common set of patients but with different attributes to collaborate. For instance, blood work results and imaging data of the same patients could be possessed by two different participating institutes~\cite{yang2019federated} but used to build a common AI model.

In this work, we consider the common horizontal FL setting in which the participating clients share model updates $\Delta W$, after training on their local data, to a central server that aggregates them, updates the global model, and sends back the updated global model to each client to continue the local training (see Fig. \ref{fig:attack_scenario}). This procedure is continued until the model converges. It is easy to see that this FL setting is equivalent to the well-known federated averaging (FedAvg) algorithm~\cite{mcmahan2017communication}. Although the raw training data is never shared in FL, there is a concern that the shared model updates between the server and clients may still leak some training data~\cite{geiping2020inverting}. While the server receives model updates from each client, a client computes model updates from the global models, consisting of updates from all clients, in different rounds of FL. We focus on the server-side gradient inversion attacks and investigate their applicability to FL in medical image classification scenarios. However, we note that the client-side gradient inversion could be treated in the same way but would involve the aggregation of more gradients from other clients. 

We investigate the risks of data leakage in gradient inversion attacks in practical FL scenarios using a real-world FL system to record the model updates. Unlike previous efforts in this area~\cite{geiping2020inverting,kaissis2021end} which assume fixed Batch Normalization (BN)~\cite{ioffe2015batch} statistics (i.e., evaluation mode), our work is the first to utilize model updates (i.e., gradients) that are generated by updating BN statistics during training, commonly known as ``training mode'' in frameworks like PyTorch or TensorFlow. Therefore, our work is directly applicable to FL use-cases where BN is used. Specifically, we propose a novel gradient inversion algorithm for estimating the running statistics of BN layers (i.e., running mean and variance) to match the gradient updates, and as a result, extract prior knowledge from intermediate feature distributions. 

To further understand the risk of gradient inversion in FL, we propose several ways to measure and visualize the potential of data leakage. In addition, we introduce a comprehensive study of realistic FL scenarios consisting of clients training with different amounts of data, batch sizes, and different differential privacy settings and utilize them to evaluate our proposed gradient inversion attack. We use our findings to make recommendations for more secure and privacy-preserving FL settings. 

Our work is inspired by recent advances in gradient inversion of deep neural networks, which also raised concerns about privacy in FL~\cite{geiping2020inverting}, and especially for healthcare applications~\cite{kaissis2021end}. One of the earliest works on this topic was Deep Gradient Leakage (DLG)~\cite{zhu2019deep}, which only relied on the information of shared gradients for their inversion attack. The main idea of their proposed algorithm is to match the gradients from synthesized and real data. However, this approach has several drawbacks which limit its use in practical FL scenarios. First, it only works for low-resolution images ($32 \times 32$ or $64 \times 64$). Second, it needs a second order optimizer (e.g., L-BFGS) for the reconstruction algorithm. Such an optimizer is computationally expensive and cannot be easily applied to scenarios with high-resolution images or larger batch sizes. The work by Geiping et al.~\cite{geiping2020inverting} attempts to address these issues and aligns itself with a more practical FL scenario by proposing a cosine similarity loss function to match the synthesized and ground truth gradients. This work uses first-order based optimizers such as Adam to make it more suitable for high resolution images and larger batch sizes. Geiping et al.~\cite{geiping2020inverting} showed also that high-resolution images could be reconstructed from model updates sent in FL from small batch sizes, with access to the model gradients for one training iteration. Most recently, Yin et al.~\cite{yin2021see} further improved the image fidelity and visual realism of gradient inverted reconstructions by matching fixed running statistics of BN layers. To effectively localize objects in the training images, they use different optimization seeds to simultaneously reconstruct several images that can be registered to a consensus image from all recovered images. Kaissis et al.~\cite{kaissis2021end} applied the framework of Geiping et al.~\cite{geiping2020inverting} to medical image classification scenarios and analyzed its applicability to reconstructing training images from FL clients. However, in their analysis, clients were trained only in the evaluation mode without updating the BN statistics, which is uncommon in modern deep learning~\cite{he2016deep,huang2017densely}. \added{A related line of work focuses on the inversion of trained models~\cite{fredrikson2015model}, e.g., ``DeepInversion''~\cite{yin2020dreaming}, but is less applicable to the particular inversion risk from gradients exchanged in FL which is the focus of our work. We summarize the key differences between these recent gradient and model inversion methods and our proposed algorithm in Table~\ref{tab:key_difference}.}

Despite remarkable insights, one common assumption by prior work is the fixed BN statistics given the most common task of single-batch gradient inversion. However, in practical FL scenarios, such statistics vary inevitably when multiple batches are jointly used by a client within one training epoch for the weight update computation. As the gradient inversion problem is a second-order optimization by nature that relies on a static underlying forward pass, any shifts in BN statistics can result in accumulated errors across batches and baffle the efficacy of prior attacks, as we show later. 
This work makes the following novel contributions:
\begin{enumerate}
\item By carefully tracing the momentum mechanism behind BN updates, this work demonstrates the first successful attempt of gradient inversion in a multi-batch scenario with updated BN statistics by introducing an FL-specific BN loss. 
\item A comprehensive analysis is provided as guidance to enhance FL security against gradient inversion attacks in both gray-scale and color image classification applications.
\item To the best of our knowledge, our work is the first to show how a gradient inversion attack might be successful when clients are sending model updates while updating their BN statistics.
\item We present new ways to measure and visualize potential data leakage in FL.
\end{enumerate}

\begin{table*}[ht]
    \centering
    \caption{\added{Key differences between recent model/gradient inversion methods \& our new inversion algorithm and their applicability to federated learning (FL) and batch normalization (BN).} \label{tab:key_difference}}
    \resizebox{\linewidth}{!}{%
    \begin{tabular}{|m{0.04\columnwidth}|m{0.6\columnwidth}|m{0.6\columnwidth}|m{0.6\columnwidth}|m{0.6\columnwidth}|}
         \hline
         & \multicolumn{1}{c|}{\textbf{DeepInversion~\cite{yin2020dreaming}}} & \multicolumn{1}{c|}{\textbf{GradInversion}~\cite{yin2021see}} & \multicolumn{1}{c|}{\textbf{Geiping}~\cite{geiping2020inverting}} & \multicolumn{1}{c|}{\textbf{Epoch-wise Gradient Inversion (Ours)}} \\\hline
         \rotatebox[origin=c]{90}{\bf \ Input \ } & Weights & Weights + Gradients & Weight change over entire epoch & Weight change over entire epoch with BN statistics update \textbf{(new)} \\\hline
         
         \rotatebox[origin=c]{90}{\bf \ Goal \ } & Data synthesis from \textit{\textbf{trained weights}} (model inversion) & Image recovery from \textit{\textbf{gradients of one batch}} & Image recovery from \textit{\textbf{epoch-wise model updates}} & Image recovery from \textit{\textbf{epoch-wise model updates}} given \textit{\textbf{multiple}} batches under \textit{\textbf{momentum}} \textbf{(new)} \\\hline
         \rotatebox[origin=c]{90}{\bf \ Setup \ } & Data-access limited applications & FL single-batch study & FL multi-batch study & FL multi-batch study with BN updates \textbf{(new)} \\\hline
         \rotatebox[origin=c]{90}{\bf \ Defense \ } & N.A. & N.A. &    N.A.    & Differential Privacy \& Data leakage quantification \textbf{(new)} \\\hline
         \rotatebox[origin=c]{90}{\bf \ Insight \ } & Weights encode dataset priors & Gradients encode private information via inversion &    Image Recovery from epoch-wise model updates in FL is possible when training \textit{\textbf{without}} updating BN statistics   & BN loss allows image recovery from model updates \textit{\textbf{with updated}} BN statistics even for epoch-wise averaging in ``high-risk'' clients \textbf{(new)} \\\hline
    \end{tabular}
    }%
\end{table*}


\section{Methods}
\label{sec:methods}

\subsection{Inversion Attacks}
\label{sec:inversion_attacks}

\begin{figure}[htbp]
    \centering
    \includegraphics[width=1.\columnwidth]{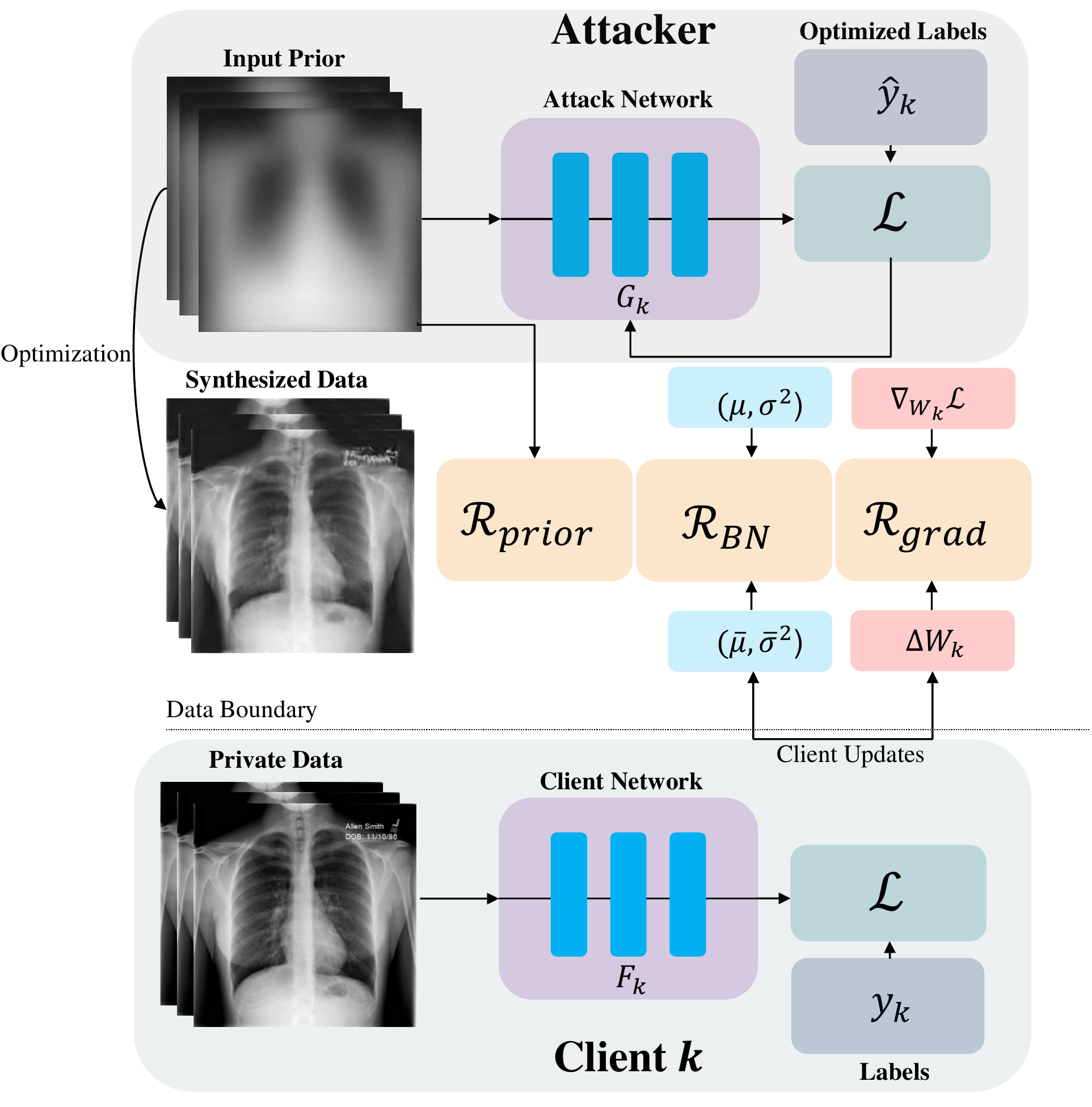}
    \caption{\textbf{Overview of the attack model.} A malicious actor (``attacker'') recovers private training data via updates sent from a client by optimizing an input prior. The objective aims to simulate the training by matching the intercepted model updates and BN statistics from a client. Additional image prior losses are added to enhance the quality of the recovered images. \label{fig:attack_overview}}
\end{figure}

Today's gradient inversion attacks follow a common principle. The main idea is to match the gradients between the trainable input data and real data. If the attack has access to the state of the global model $W$ and a client's model update $\Delta W$, it can try to optimize the input image (usually randomly initialized or using a prior image) to produce a gradient that matches the observed model updates (see Figure \ref{fig:attack_overview}).

\subsection{Practical Gradient Inversion Attack}
\label{sec:practical_attack}

Although previous efforts have shown the possibility of reconstructing training data by matching the gradients and optimizing the randomly initialized input images, these attacks have been conducted in contrived settings. Prior methods simplified the attack scenario by sending model updates with fixed BN statistics. However, in realistic FL scenarios, these statistics change during training with multiple batches in an epoch before sending the updates to the server. Since the gradient inversion is formulated as a second-order optimization task, accumulation of error due to discrepancies in the estimation of BN statistics leads to sub-optimal solutions. As a result, matching the gradients in practical use cases is extremely difficult, if not impossible, when taking these considerations into account, especially when the training takes more than one local iteration. In this work, we study a real-world attack scenario in which the intercepted model updates from the clients are sent while updating the BN statistics.

In networks that leverage BN layers~\cite{ioffe2015batch} in their architecture, during training, mini-batches of input data $x$ are first normalized by the batch-wise mean $\mu$ and variance $\sigma^{2}$ and transformed by an affinity mapping with trainable values $\beta$ and $\gamma$ according to $\frac{x-\mu(x)}{\sqrt{\sigma^{2}(x)+\epsilon}}  \gamma+\beta$. Running statistics of $\mu$ and $\sigma$ are updated during training according to the momentum update rule~\cite{ioffe2015batch}.
These BN layers contain a strong prior in the forms of tracked running statistics of the training data and can be effectively leveraged in gradient-based attack scenarios. During evaluation (i.e., inference), the estimated running mean and variance are used in lieu of computing on-the-fly batch-wise statistics. However, previous efforts~\cite{zhu2019deep,geiping2020inverting,kaissis2021end} have not taken into account the effect of running statistics in the BN layers and only utilized gradient updates that were sent by assuming fixed BN statistics. As a result, matching the gradient updates that are sent during the training phase is not possible since the running statistics from BN layers affect the gradient computation. In our attack scenario, we consider the server as ``honest but curious''~\cite{bonawitz2017practical}. Hence all global models, configuration files, local model updates and training BN statistics that exist on the server or are exchanged between the server and client can be considered as compromised for a potential gradient inversion attack. 

Figure \ref{fig:attack_overview} illustrates an overview of our attack model. We formally define the attack scenario for one server-client configuration which can be extended to all participating clients in FL. Consider client $k$ having local model $F_{k}$ with weights $W_{k}$, the model updates $\Delta W_{k}$ are computed with respect to mini-batches from the local training images $x_{k}$ and their corresponding labels $y_{k}$. In order to reconstruct the private data, we formulate an optimization scheme wherein trainable input images $\hat{x}_{k}$ and the corresponding labels $\hat{y}_{k}$ are employed to match the client’s updates by minimizing
\begin{equation}
\underset{\hat{x}_{k}}{\text{argmin }}{ {\mathcal{L_\text{grad}}(\hat{x}_{k}; \Delta W_{k}) + \mathcal{L_\text{BN}}(\hat{x}_{k})+ \mathcal{L_\text{prior}}(\hat{x}_{k})}}.
\label{eqn:main_error}
\end{equation}
\noindent
Here, $\mathcal{L_\text{grad}}$ is the gradient matching loss between the intercepted gradients and computed gradients using $\hat{x}_{k}$ and $\hat{y}_{k}$ in each layer $l$ as in
\begin{equation}
\mathcal{L_\text{grad}}(\hat{x}_{k}; \Delta W_{k}) 
    = \sum_{l} || \nabla_{W^{(l)}_{k}}\mathcal{L}(\hat{x}, \hat{y}) - \Delta W^{(l)}_{k}||_2.
\label{eqn:lgrad}
\end{equation}
\noindent
Where $F_{k}$ and $G_{k}$ denote the client and the attack network, respectively. We initialize the attack network $G_{k}$, used for computing the matching gradient, by the global weights on the server and update it with the model weight differences $\Delta W^{(l)}_{k}$ sent from client $k$. In addition, we update the intercepted BN updates from the client by computing the running statistics with batch-wise mean $\mu_{k}$ and variance $\sigma_{k}$ and initialize the BN layers in the attack network with these updated statistics. For this purpose, we estimate the BN running statistics of each mini-batch by
\begin{equation}
\bar{\mu}_\text{new}=\left(1-\eta_\text{BN}\right) \bar{\mu}+\eta_\text{BN} \mu_\text{new},
\label{eqn:bn_1}
\end{equation}
\noindent
\begin{equation}
{\bar{\sigma}^{2}}_\text{new}=\left(1-\eta_\text{BN}\right) {\bar{\sigma}^{2}}+\eta_\text{BN} {\sigma^{2}}_\text{new}.
\label{eqn:bn_2}
\end{equation}
\noindent
Here, $\bar{\mu}$ and $\bar{\sigma}$ denote the running mean and variance, respectively, and $\eta_\text{BN}$ is set to $0.1$. These considerations ensure that the attack model exactly simulates the client training procedure by matching the BN statistics during training, hence making it applicable to practical FL scenarios. Note, that if the number of local iterations increases, the estimation of running statistics becomes less accurate and hence it becomes harder to properly match the gradients. This effect is illustrated in Fig. \ref{fig:reconstructions} and Fig. \ref{fig:ablations} where we show the effect of the number of training images, batch size, and local iterations on the gradient inversion attack. 

$\mathcal{L_\text{BN}}$ is also employed to match the distributions of BN statistics obtained from the trainable input $\hat{x}_{k}$ and intercepted updates from the client $k$ in each layer $l$ and computed according to
\begin{equation}
    \mathcal{L_\text{BN}}(\hat{x}_{k})
    = \sum_{l}|| \ {{\mu}_{l}}(\hat{x}_{k}) - \bar{\mu}_{l} ||_2 + \sum_{l}|| \ {{\sigma}_{l}^{2}}(\hat{x}_{k}) - {\bar{\sigma}_{l}}^{2} ||_2.
\label{eqn:bn_loss}
\end{equation}
\noindent
We account for image prior losses by minimizing $\mathcal{L_\text{prior}}$, which includes total variation and $l_{2}$ norm losses~\cite{yin2021see}. Similar to ~\cite{zhu2019deep}, we jointly optimize the corresponding labels from a client's data by using a trainable label $\hat{y}_{k}$ and minimize a loss function such as cross-entropy that is used for supervising the classification networks. For initializing the trainable parameters, and without loss of generality, we initialize $\hat{x}_{k}$ from a domain-specific prior (e.g., chest X-ray images) for faster convergence and avoiding sub-optimal solutions. At the same time, the attacked labels $\hat{y}_{k}$ are randomly initialized from a uniform distribution in the interval $0$ to $1$. 

\subsection{Differential Privacy (DP)}
\label{sec:dp}

Most commonly, DP mechanisms add some form of calibrated noise to the data in order to obfuscate the contributions of individual data entries, i.e., patients in healthcare applications~\cite{dwork2006calibrating}. There are many variations of how one can add this noise, like the Gaussian mechanism or Laplacian mechanism~\cite{yang2020local}. Other, more sophisticated methods like the sparse vector technique (SVT) combine the addition of noise with value clipping and/or removal of parts of the data~\cite{lyu2016understanding}. DP has been successfully integrated into the training of deep neural networks~\cite{abadi2016deep,wu2019p3sgd,ziller2021medical} and used for FL applications~\cite{kaissis2021end,li2019privacy,liang2020differentially,rodriguez2020federated,adnan2022federated}. DP can make strong guarantees about the theoretical effectiveness of protecting against individual privacy loss but is difficult to tune to keep the model accuracy at an acceptable level~\cite{bagdasaryan2019differential}.

\replaced{
To reduce the risk of server-side gradient inversion in FL, we must protect the outgoing message (model update), not necessarily the local training itself. Therefore, FL applications of DP often follow an approach that adds noise to information (i.e., gradients) shared between participants of FL~\cite{li2019privacy}. This approach has the added advantage that the differential privacy filter applied to model updates does not depend on the optimizer used for local training, and implementations of DP can be independent of the local learning algorithm.

An alternative approach for general deep learning applications is ``differentially private SGD'' (DP-SGD), which uses an optimizer that applies noise at each iteration of the training~\cite{abadi2016deep}. DP-SGD could be helpful in both centralized and FL scenarios as it reduces the risk of the model memorizing sensitive data~\cite{song2020systematic,shokri2015privacy}.
}{
In deep learning, ``differentially private SGD'' (DP-SGD) has been proposed as an optimizer that applies noise at each iteration of the training~\mbox{\cite{abadi2016deep}}. In FL, we only need to protect the outgoing message (model update), not the local training itself. Therefore, FL applications of DP often follow an approach that adds noise to information (i.e., gradients) shared between participants of FL~\mbox{\cite{li2019privacy}}. This has the added advantage that the differential privacy filter applied to model updates does not depend on the optimizer used for local training and implementations of DP can be independent of the local learning algorithm.
}

Our work intentionally utilizes a simple Gaussian mechanism to explore DP and to establish an upper bound for successful inversion attacks in the FL scenario. Our intention is to illustrate how simply adding Gaussian noise to the model updates can already reduce the success of the inversion attack but at the cost of decreasing model performance.

At each round of FL, we compute the $q$th percentile of the absolute model update values and multiply this value by $\sigma_0$, a tuneable hyperparameter that can be specified by the user and controls the overall amount of noise added to the model updates. The resulting $\sigma$ is then used to sample from a Gaussian distribution $N\left(0,\sigma\right)$ with zero-mean. Therefore, the model updates become $\Delta W = \Delta W + N\left(0, \sigma \right)$, where $\sigma=percentile\left(\Delta W, q\right)\sigma_0$, before they are sent to the central server for aggregation. In our experiments, we use $q=95$, which was chosen to reduce the impact of outliers in the gradient computation, which in turn would cause too high an amount of noise and hinder the model convergence.

As this DP protocol is applied locally on each client before the updates are sent to the server, this approach is commonly referred to as local DP~\cite{yang2020local}. In scenarios where only the protection against model inversion after FL training is of concern (e.g., when employing cryptographic techniques such as homomorphic encryption~\cite{zhang2020batchcrypt} to make the FL aggregation step itself secure against inversion), ``global'' or ``central'' DP could be utilized to protect against model inversion after training is completed~\cite{naseri2020local}.

\added{We also evaluate the potential data leakage when clients apply DP-SGD as their local training method. DP-SGD clips the gradients during local training based on the maximum $l_{2}$ norm of the gradients and a noise multiplier $\sigma_\mathrm{dp}$ relative to the clipping norm value $C_\mathrm{dp}$~\cite{mcmahan2018general}, see Fig.~\ref{fig:dp-sgd}.} 
Future work could explore \replaced{alternative}{more sophisticated} DP or gradient perturbation methods~\cite{li2019privacy,sun2021soteria} and how they impact the quality of inversion attacks and their impact on subsequent model inversion attacks.

\subsection{Image Identifiability Precision}
\label{sec:iip}
In this work, we use image identifiability precision (IIP) ~\cite{yin2021see}, which is designed for measuring the image-specific features between the reconstructed images and original training data. In its original formulation, the metric is computed for a fixed number of randomly selected images (e.g., 256) by employing the gradient-inversion attacks for each batch of data. The deep feature embeddings of reconstructions, as well as the entire pool of training images are computed using a pretrained network (e.g., ResNet-18 trained on ImageNet). Furthermore, with a clustering algorithm (e.g., K-Nearest Neighbors), the closest training image in the feature embedding space (as measured by cosine similarity) of each reconstructed image is selected and fetched back as an exact match if it matches the original image. The IIP score is then computed in as $\mathrm{IIP}=\frac{m}{N}$, where $m$ is the number of unique exact matches according to the described procedure. Here, $N$ denotes the size of the training pool (e.g., 256). In this work, we use a modified version of the IIP metric adjusted to the FL scenario in which the IIP score is computed for the entire training and validation images of all clients as opposed to the randomly selected samples. A reconstruction is counted as an exact match if the closest match originates from the attacked client. As a result, a diverse population of training images can be taken into consideration for IIP calculations to quantify data leakage. 
\section{Results}
\label{sec:results}
\subsection{FL Scenario}
\label{sec:fl_scenario}
We simulate eight clients using different numbers of training images and local batch sizes in a cross-silo FL scenario with horizontal data partitioning for a chest X-ray image classification task. Each client trains an ImageNet-pretrained~\cite{deng2009imagenet} ResNet-18~\cite{he2016deep} for one epoch \added{per round} using a stochastic gradient descent (SGD) optimizer without momentum and a step-wise learning rate decay \added{by a factor of 0.1 at every 40 rounds} before sending model updates to the server \added{for a total of 100 rounds.}
We study the gradient inversion attacks in both gray-scale and color image classification applications using two datasets, A and B.
Dataset A consists of publicly available chest X-ray images from normal patients and patients who tested positive for COVID-19~\cite{chowdhury2020can,rahman2021exploring}. As input prior, we compute a mean image from a different publicly available dataset, namely the ChestX-ray14 dataset~\cite{wang2017chestx}.
Dataset B consists of publicly available color fundus images from the Rotterdam EyePACS AIROGS train set~\cite{de_vente_coen_2021_5793241}. The input prior for experiments using dataset B was generated by computing the mean image over images available through the ODIR-2019 challenge.
The data used for training, validation, and testing data for both datasets is available online\footnote{
Dataset A (Chest X-ray):
\href{https://www.kaggle.com/tawsifurrahman/covid19-radiography-database}{FL data}
\href{https://nihcc.app.box.com/v/ChestXray-NIHCC}{prior data};
Dataset B (Fundus color images):
\href{https://zenodo.org/record/5793241\#.YxpjcNLMKEB}{FL data}
\href{https://odir2019.grand-challenge.org}{prior data}
}.
Table~\ref{tab:dataset} shows a summary of the local training batch size and number of training images that were used for each client. Each client has its own validation set of 200 cases that will be used to select the best global model based on the average validation accuracy. The selected global model is finally evaluated on the testing set. Except for client 9, each client training set contains the same number of normal and COVID-positive cases in order to simulate clients with balanced class distributions. Client 9 shares its only training image (a normal case) and its validation set with client 1. As a result, client 9 is at the highest risk of data leakage as it is sending model updates based on a single image and a batch size of one and will be referred to as the ``high-risk client''. We intentionally utilize relatively small batch sizes in order to study the effect of our new inversion attack on clients with varying amounts of training data (and therefore, local training iterations)\footnote{All experiments are using \href{https://ngc.nvidia.com/catalog/containers/nvidia:clara-train-sdk}{Clara Train SDK 4.0} for local model training and evaluation. 
\href{https://developer.nvidia.com/flare}{NVIDIA FLARE}~\cite{Roth_NVIDIA_FLARE_Federated_2022} is used to record model updates in a simulated FL setting. The inversion software is implemented in \href{https://pytorch.org/}{PyTorch} and available at \url{https://github.com/NVlabs/DeepInversion}. \added{We use a PyTorch version of \href{https://github.com/srxzr/DPSGD}{\textit{tf.privacy.DPGradientDescentGaussianOptimizer}} to implement DP-SGD.}}. Therefore, all clients investigated in this study could be considered of higher risk than in real-world FL scenarios where clients use larger amounts of training data and batch sizes (see also the discussion in Section \ref{sec:discussion}).

\begin{table}[htbp]
\centering
\caption{Dataset used for simulating federated learning. Note, the number of training and validation cases are the same across datasets A and B. \label{tab:dataset}}
\begin{tabular}{p{0.16\linewidth}p{0.14\linewidth}p{0.14\linewidth}p{0.16\linewidth}}
\toprule
\textbf{client}        & \textbf{batch size} & \textbf{\# training} & \textbf{\# validation}\\
\hline
client 1               & 4                   & 8                            & 200                            \\
client 2               & 4                   & 32                           & 200                            \\
client 3               & 4                   & 128                          & 200                            \\
client 4               & 4                   & 512                          & 200                            \\
client 5               & 8                   & 8                            & 200                            \\
client 6               & 8                   & 32                           & 200                            \\
client 7               & 8                   & 128                          & 200                            \\
client 8               & 8                   & 512                          & 200                            \\
\hline
client 9\\ (``high-risk'') & 1                   & 1                            & 200                          \\
\hline
\textbf{sum:}    &                            & 1360                         & 1600                           \\             
\bottomrule
\textbf{\# testing A:} & 1382 & & \\
\textbf{\# testing B:} & 10145 & & \\
\bottomrule
\end{tabular}
\end{table}

\subsection{Limitations of Previous Attack Scenarios}
\label{sec:limitation_prior_works}

\begin{figure*}[htbp]
\newcommand{\figwidth}{.16\textwidth}
\newcommand{\figheight}{.95\textwidth}
\centering
\begin{subfigure}{\figwidth}
  \centering
  \includegraphics[height=\figheight]{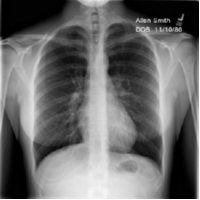}
  \caption{\scriptsize \\ \centering Original}
  \label{fig:attack_success_a}
\end{subfigure}
\begin{subfigure}{\figwidth}
  \centering
  \includegraphics[height=\figheight]{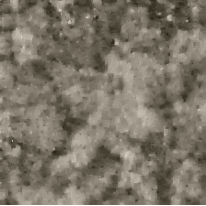}
  \caption{\scriptsize\centering  w/o BN loss,\hspace{\textwidth} w global ckpt \cite{geiping2020inverting}}
  \label{fig:attack_success_b}
\end{subfigure}
\begin{subfigure}{\figwidth}
  \centering
  \includegraphics[height=\figheight]{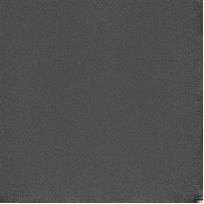}
  \caption{\centering\scriptsize w BN loss,\hspace{\textwidth} w/o global ckpt}
  \label{fig:attack_success_c}
\end{subfigure} 
\begin{subfigure}{\figwidth}
  \centering
  \includegraphics[height=\figheight]{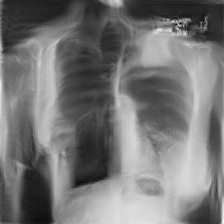}
  \caption{\centering\scriptsize w BN loss,\hspace{\textwidth} w global ckpt (no prior)}
  \label{fig:attack_success_d}
\end{subfigure} \\
\begin{subfigure}{\figwidth}
  \centering
  \includegraphics[height=\figheight]{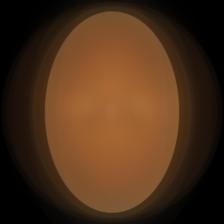}
  \caption{\added{\scriptsize \\ \centering Fundus prior}}
  \label{fig:attack_fundus_prior}
\end{subfigure}
\begin{subfigure}{\figwidth}
  \centering
  \includegraphics[height=\figheight]{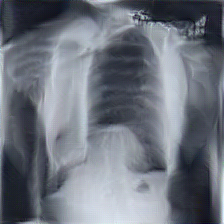}
  \caption{\centering\scriptsize \added{w BN loss,}\hspace{\textwidth} \added{w global ckpt (wr. prior)}}
  \label{fig:attack_success_wrongprior}
\end{subfigure}
\begin{subfigure}{\figwidth}
  \centering
  \includegraphics[height=\figheight]{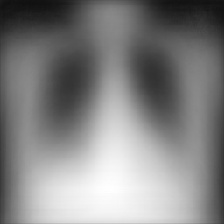}
  \caption{\scriptsize \\ \centering \added{CXR} prior}
  \label{fig:attack_success_prior_img}
\end{subfigure}
\begin{subfigure}{\figwidth}
  \centering
  \includegraphics[height=\figheight]{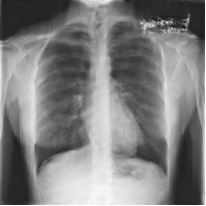}
  \caption{\centering\scriptsize \textbf{Ours}: w BN loss,\hspace{\textwidth} w global ckpt (prior)}
  \label{fig:attack_success_noprior}
\end{subfigure}
    \caption{\textbf{Success factors of the inversion attack.} 
    Here, the ``high-risk'' client 9 sends updates based on one iteration and one image in round 10 of an FL experiment. The original training image is shown for comparison (a). 
    \replaced{
    The inversion technique by Geiping et al.~\cite{geiping2020inverting} does not estimate updated BN statistics during training and therefore cannot handle clients with updated BN layers in their architectures (b). The attack also fails if not initialized with the current global model state/checkpoint (ckpt) (c).
    Both the current global checkpoint and proper accounting for updated BN statistics during training are necessary for the attack to succeed. However, the reconstruction lacks ``anatomical coherence'' without using a prior (d). Using the incorrect prior, e.g., a fundus prior image (e) as used in later experiments with RGB images, and not constraining the reconstruction to be only one channel (gray-scale), the anatomical coherence and color similarity are further reduced (f). Using the correct CXR prior (g) and forcing a gray-scale reconstruction, we can achieve the highest reconstruction qualitatively (h).
    }{
    The prior image used to initialize the attack is shown in (b). The inversion technique by Geiping et al.\mbox{~\cite{geiping2020inverting}} does not estimate updated BN statistics during training and therefore cannot handle clients with updated BN layers in their architectures (c). The attack also fails if not initialized with the current global model state/checkpoint (ckpt) (d). Both the current global checkpoint and proper accounting for updated BN statistics during training are necessary for the attack to succeed (e) and further improves when using the image prior (f). The patient’s name and date of birth in the original image are randomly generated to illustrate the difficulty of inverting detailed textual information by the gradient inversion attack.}
    \label{fig:attack_success}}
\end{figure*}

We utilize an extended version of the GradInversion algorithm~\cite{yin2021see} to invert the training images of each client separately based on its model updates sent during FL. For details on the inversion attack, see Section \ref{sec:practical_attack}. Prior works~\cite{geiping2020inverting,kaissis2021end} did not consider clients updating the statistics of BN layers which are commonly used in modern classification models with deep convolutional neural networks (CNNs)~\cite{ioffe2015batch,he2016deep,huang2017densely}. Therefore, the clients in these efforts only use updated model weights while executing the training with fixed BN statistics. Here, we show that when the attacked clients update BN layers during training, an additional BN loss is essential for the attack to succeed. The running mean and variance of the BN layers need to be updated during the optimization of the inversion attack as these statistics inevitably change during training. Since a gradient inversion attack is a second-order optimization problem, the accumulated errors in estimating the BN statistics by assuming fixed values hinder successful reconstructions of training data (see Fig. \ref{fig:attack_success_b}). In addition, it is critical to initialize the network of inversion attack with the same global model weights that are used to initialize the client’s network in a particular FL round to ensure the exact simulation of local training steps within the attack (see Fig. \ref{fig:attack_success_c}).   
The exploitation of BN statistics due to our novel BN loss causes the quality of reconstructions to remain relatively constant throughout the training procedure as can be observed in Fig.~\ref{fig:training_stage}.

\subsection{Measuring Data Leakage by Inverting Gradients}
\label{sec:measure_leakage}

\begin{figure*}[htbp]
\centering
\begin{subfigure}{.8\textwidth}
  \centering
  \includegraphics[width=.9\linewidth]{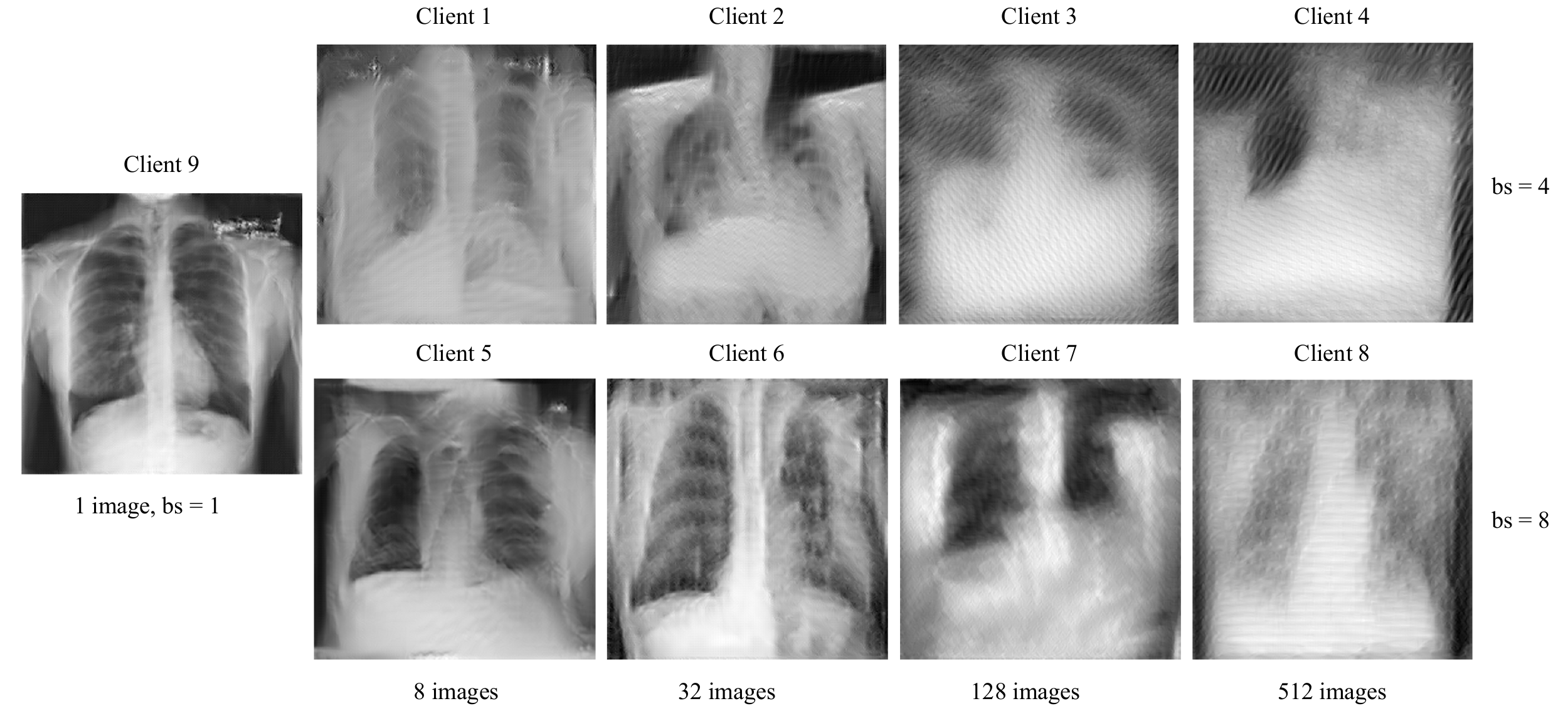}
  \caption{Round 10\label{fig:reconstructions_a}}
\end{subfigure}\\
\begin{subfigure}{.8\textwidth}
  \centering
  \includegraphics[width=.9\linewidth]{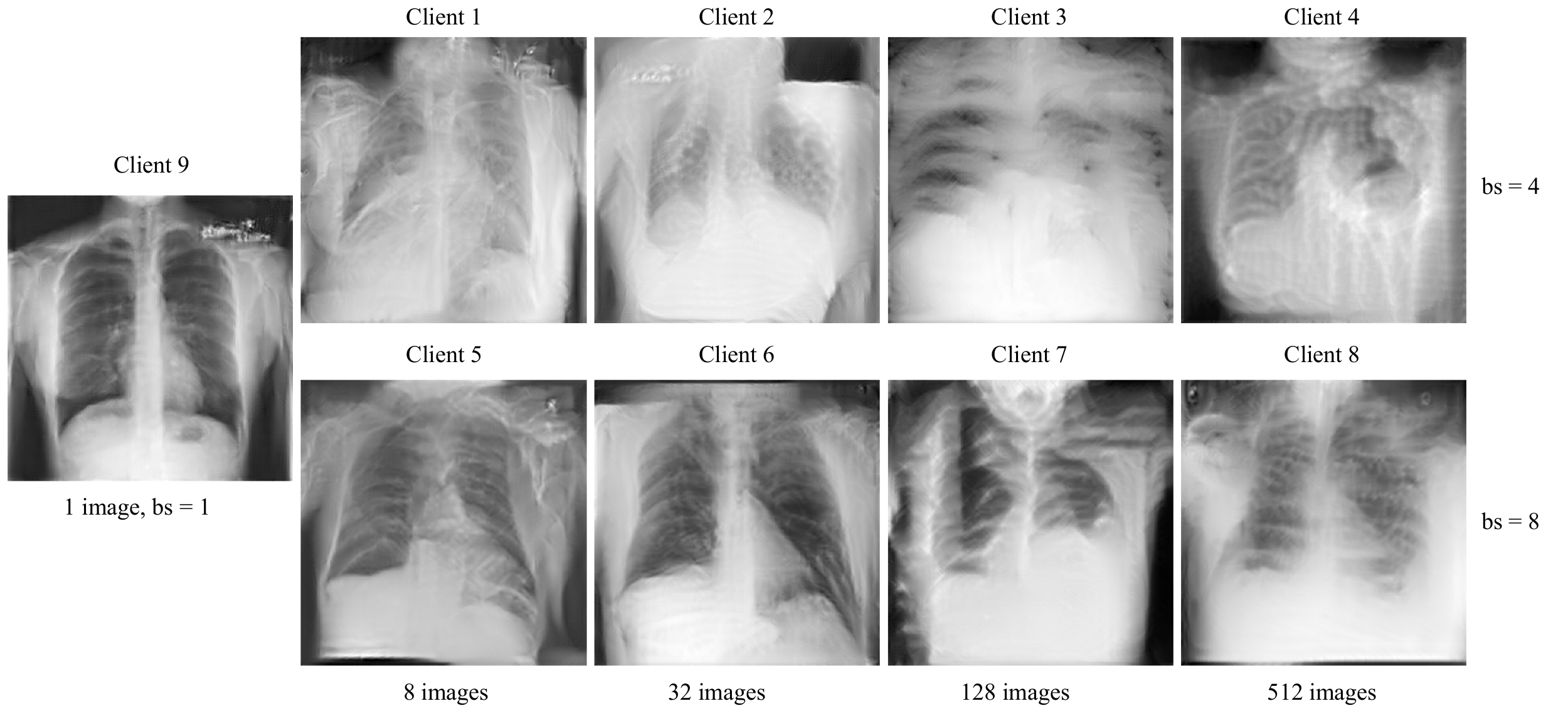}
  \caption{Round 90\label{fig:reconstructions_b}}
\end{subfigure}
    \caption{\textbf{Inversions from different FL rounds.} Effect of the number of training images, batch size, and local iterations on the inversion attack early on in FL training (a) and late in training (b). \label{fig:reconstructions}}
\end{figure*}

\begin{figure*}[htbp]
\newcommand{\figwidth}{1.0\textwidth}
\newcommand{\figheight}{.12\textwidth}  
\newcommand{\graphheight}{.15\textwidth}  
\newcommand{\hspacing}{.05\textwidth}  
\centering
\begin{subfigure}{\figwidth}
  \textbf{\hspace{.04\textwidth}Original\hspace{.065\textwidth} Round 0\hspace{\hspacing} Round 10\hspace{\hspacing} Round 20\hspace{\hspacing} Round 50\hspace{\hspacing} Round 90\hspace{.045\textwidth} Gradient norm}\par\medskip
  \centering
  \includegraphics[height=\figheight]{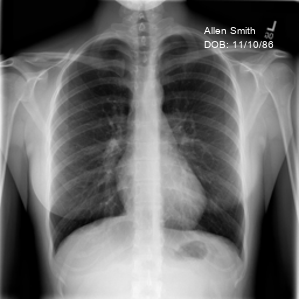}
  \includegraphics[height=\figheight]{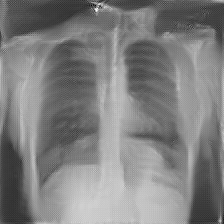}
  \includegraphics[height=\figheight]{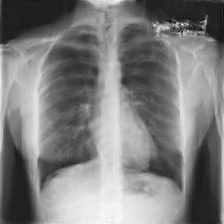}
  \includegraphics[height=\figheight]{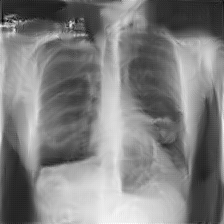}
  \includegraphics[height=\figheight]{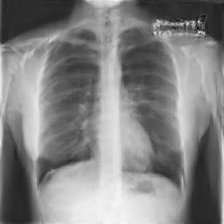}
  \includegraphics[height=\figheight]{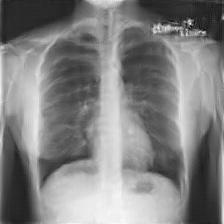}
  \includegraphics[height=\graphheight]{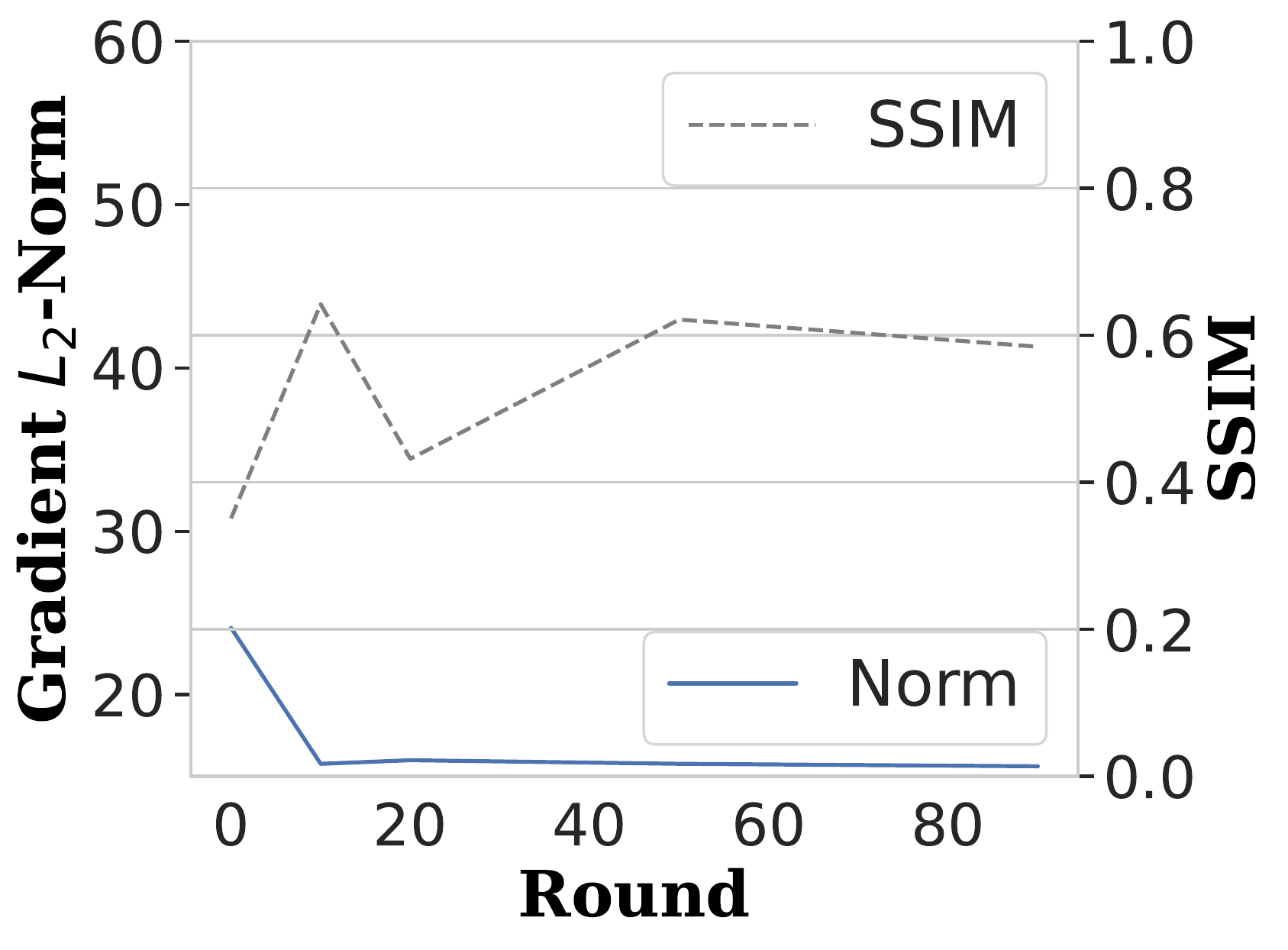}
  \caption{\scriptsize\centering Chest X-ray - ImageNet pretrained}
  \label{fig:training_stage_cxr}
\end{subfigure}
\begin{subfigure}{\figwidth}
  \centering
  \includegraphics[height=\figheight]{figs/training_stage/cxr_original/Normal-4085.png}
  \includegraphics[height=\figheight]{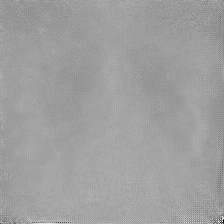}
  \includegraphics[height=\figheight]{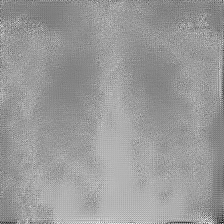}
  \includegraphics[height=\figheight]{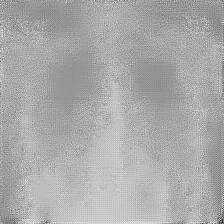}
  \includegraphics[height=\figheight]{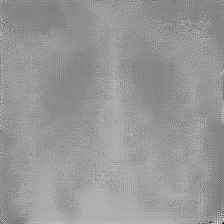}
  \includegraphics[height=\figheight]{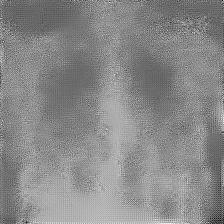}
  \includegraphics[height=\graphheight]{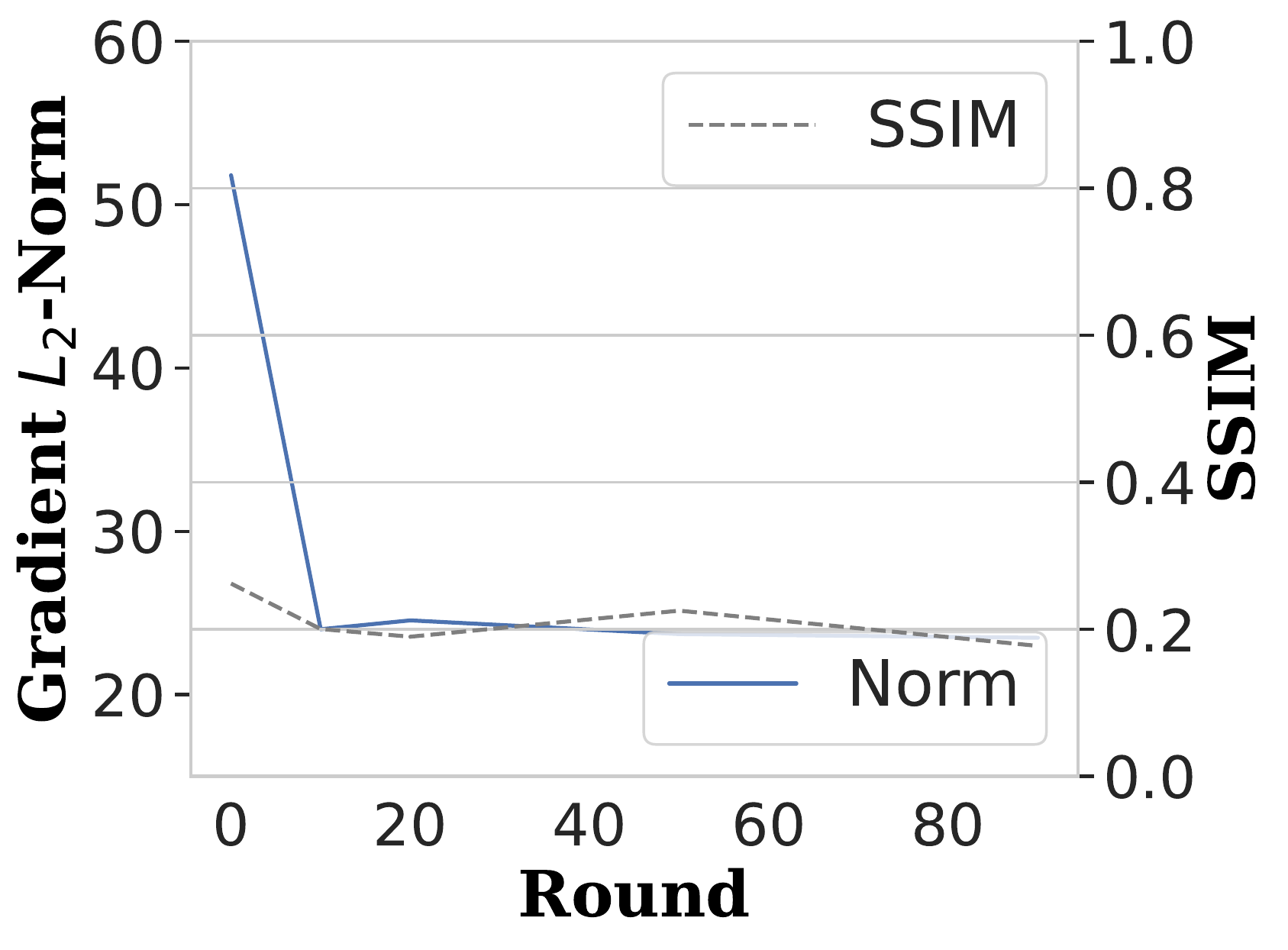}
  \caption{\scriptsize\centering Chest X-ray - From scratch}
  \label{fig:training_stage_cxr_scratch}
\end{subfigure}
\begin{subfigure}{\figwidth}
  \centering
  \includegraphics[height=\figheight]{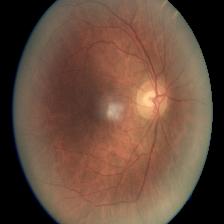}
  \includegraphics[height=\figheight]{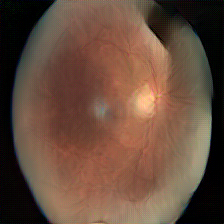}
  \includegraphics[height=\figheight]{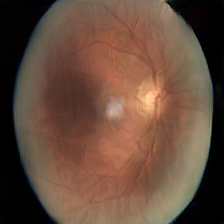}
  \includegraphics[height=\figheight]{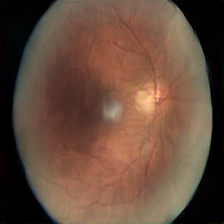}
  \includegraphics[height=\figheight]{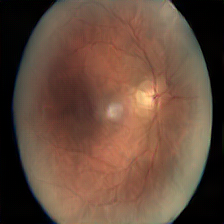}
  \includegraphics[height=\figheight]{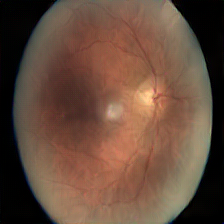}
  \includegraphics[height=\graphheight]{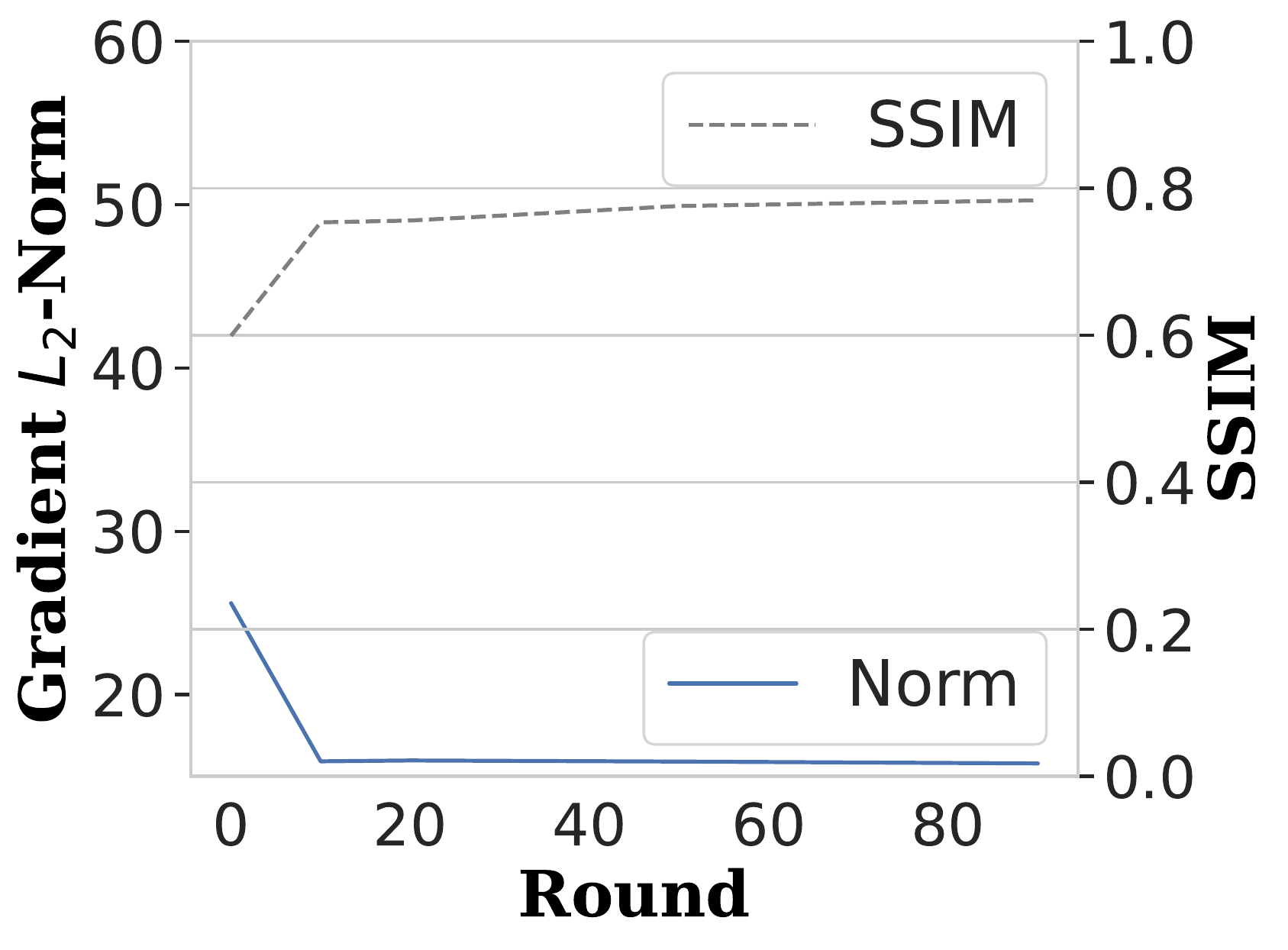}
  \caption{\scriptsize\centering Fundus images - ImageNet pretrained}
  \label{fig:training_stage_fundus}
\end{subfigure}
    \caption{\textbf{Impact of the training phase on gradient inversion.} We show inversions based on updates from the ``high-risk'' client 9 in both chest X-ray (a, b) and fundus image applications (c). As the models becomes more conditioned on the task at hand, the updates tend to leak for information. Starting training from scratch (b) is less likely to result in recognizable inversions. On the right, we show a comparison of the gradient $l_2$-norm and SSIM with respect to the current FL round. \added{Note that a step-wise learning rate decay was applied at every 40 rounds (factor of 0.1) of FL training.} \label{fig:training_stage}}
\end{figure*}

Assuming the attacker has access to the full information needed for a successful data recovery as shown in Fig. \ref{fig:attack_success_d}, we attempt to invert any model updates coming from any of the clients in different rounds of FL. Figure \ref{fig:reconstructions} illustrates the effect of the number of training images, batch size, and local iterations on the success of a gradient inversion attack in an earlier or later round of FL. A larger number of local training iterations in clients with larger training sets detrimentally impacts the quality of reconstructions due to accumulated errors in estimating the updated BN statistics. In addition, reconstructions from later rounds of FL (e.g., round 90 of 100) have improved image quality in terms of image fidelity and capturing the details of anatomical structures. See also Fig.~\ref{fig:training_stage}, where we show both the effect of the training stage on the high-risk client 9 for chest X-ray and fundus datasets. This observation might be attributed to the global model having learned more effective feature representations during the later stages of the FL training and outliers or individual images in those training sets that cause higher losses in later stages of training and therefore contribute to more recognizable reconstructions. This effect is further illustrated in Fig.~\ref{fig:training_stage} where we also show the reconstruction quality of when using a less-well training network, i.e. a network trained from scratch.



The effect of training batch sizes and local number of iterations is further quantified in Fig.~\ref{fig:vary_batchsize} and Fig.~\ref{fig:vary_iterations}, respectively. Here, we utilize the global model from round 90 for the attack and train on a client with increasing dataset sizes. In the first setting, we set the number of local iterations to one but keep increasing the batch sizes. In the second setting, we set the batch size to one but increase the local number of iterations. For each setting, we add new images to the client’s training set while measuring the impact of batch size and local number of iterations by using the same original image for reference. As shown in Fig.~\ref{fig:ablations}, increasing both the batch size and local number of iterations are detrimental to the inversion attack. In particular, increasing the local number of iterations negatively impacts the reconstructions as the estimation of BN statistics becomes less accurate when the client is iterating over different local training images (see Section~\ref{sec:practical_attack}). To quantify the data leakage between the original image and the reconstructions, we use Structural SIMilarity index (SSIM), a pixel-wise metric for measuring image similarity in a more intuitive and interpretable manner compared to other commonly used metrics such as root-mean-squared error or peak signal-to-noise ratio~\cite{wang2004image}.

\subsection{Mitigating Inversion Through Differential Privacy}
\label{sec:mitigation_dp}
\begin{figure*}[htbp]
\begin{subfigure}{.49\textwidth}
  \centering
  \includegraphics[width=1.\linewidth]{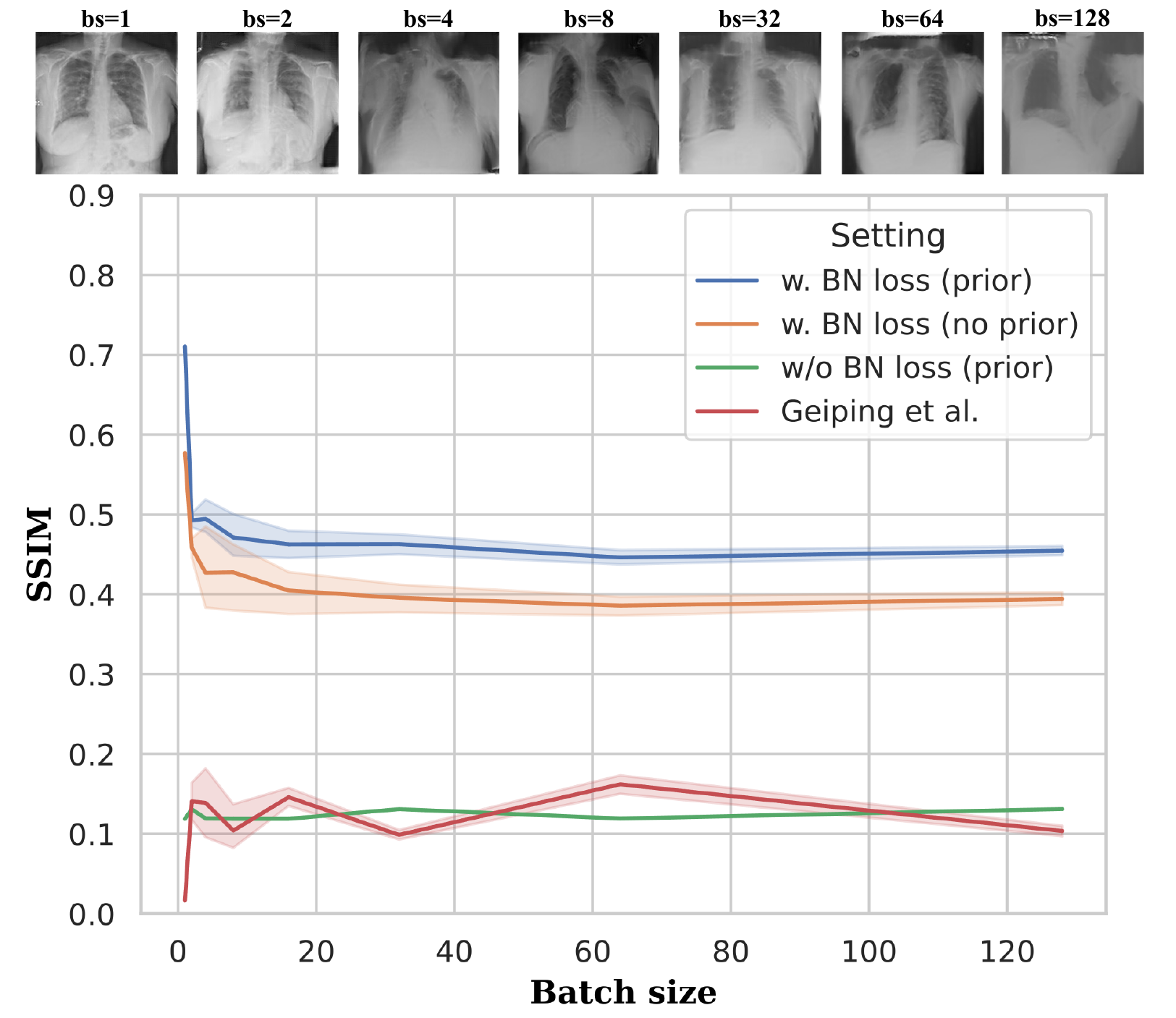}
  \caption{}
  \label{fig:vary_batchsize}
\end{subfigure}
\begin{subfigure}{.49\textwidth}
  \centering
  \includegraphics[width=1.\linewidth]{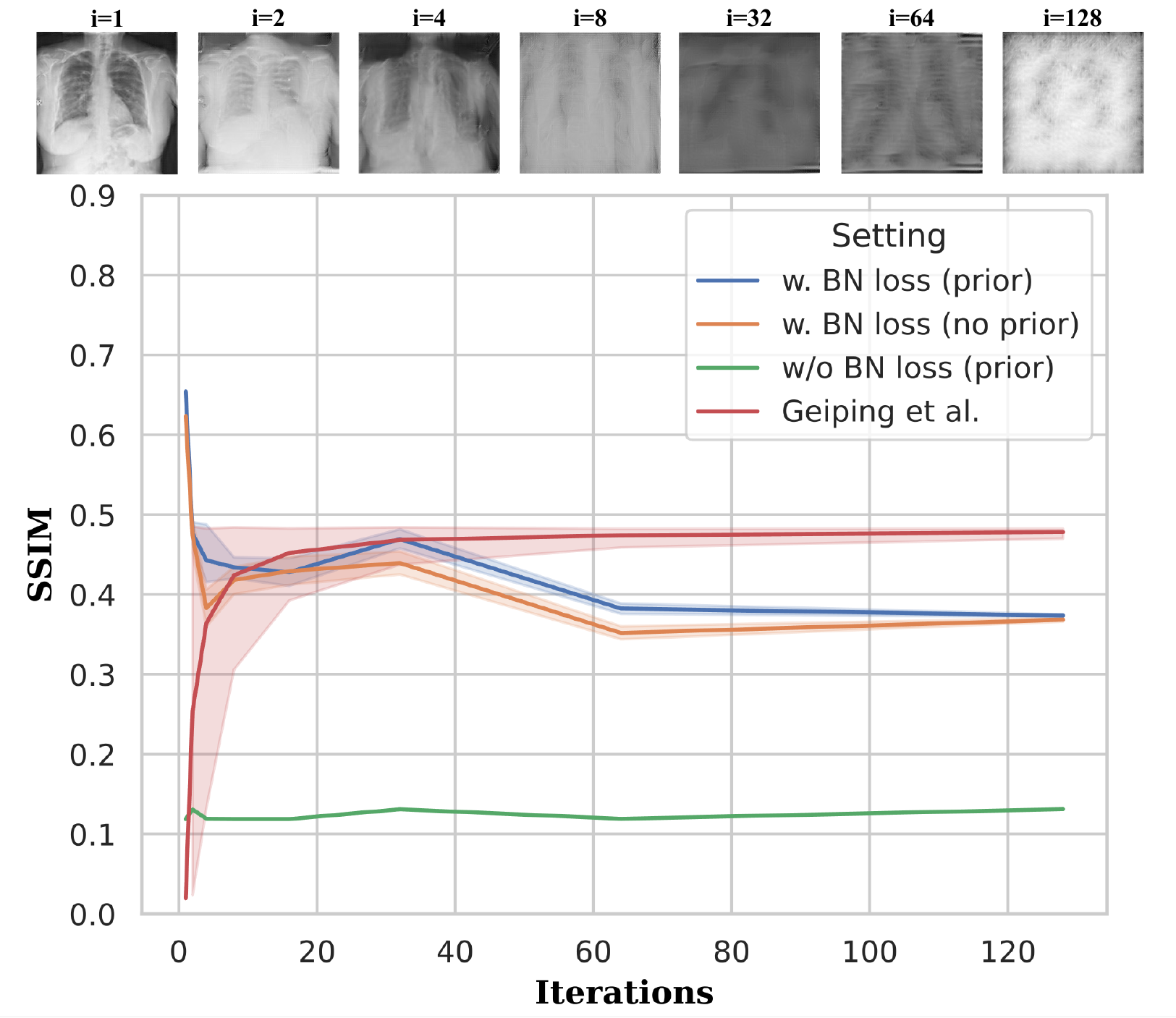}
  \caption{}
  \label{fig:vary_iterations}
\end{subfigure} \\
\centering
\begin{subfigure}{.49\textwidth}
  \centering
  \includegraphics[width=1.\linewidth]{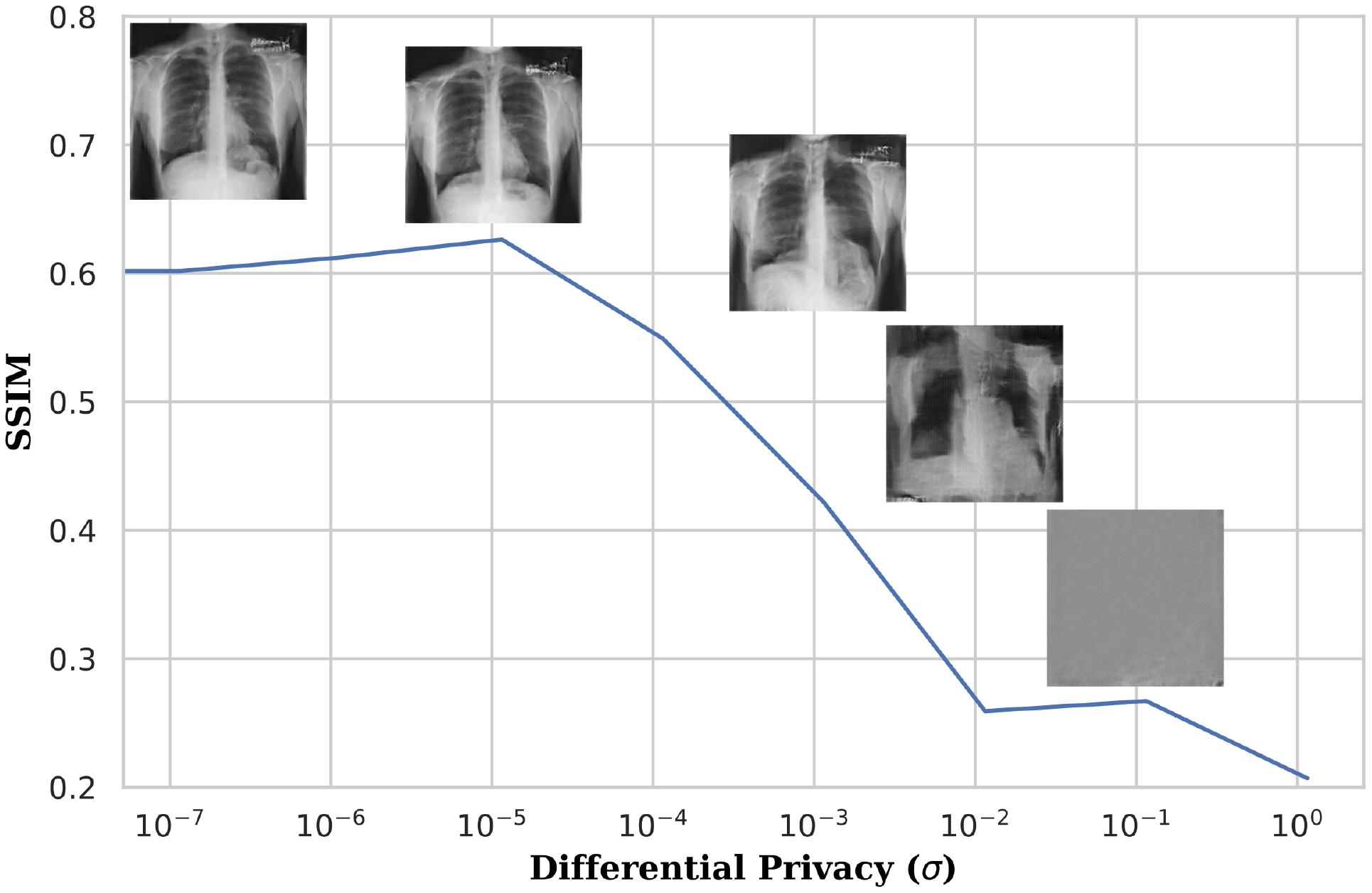}
  \caption{}
  \label{fig:vary_dp}
\end{subfigure}
    \caption{\textbf{Quantifying data leakage.} The impact of batch size (a) and local number of iterations (b) to the success of the inversion attack for different attacks either using the proposed BN loss, the prior, or both. We also show examples the inversions for different batch sizes and iterations when using both BN loss and prior (blue). The baseline inversion attack by Geiping et al.~\cite{geiping2020inverting} is shown for reference. 95\% confidence interval bands are shown for each client. Note, that a SSIM value of $\sim$0.5 still indicates no recognizable inversion result. (c) The ``high-risk'' client 9 is shown with varying amounts of noise added to the model updates before they are sent to the server. The SSIM value of the reconstructed inversions are shown for reference. \label{fig:ablations}}
\end{figure*}

A straightforward approach to avoid data recovery by a server-side gradient inversion attack is the addition of random noise to the model updates before they are sent to the server~\cite{li2019privacy}. Here, we explore a simple DP protocol that adds calibrated Gaussian noise, with zero mean, based on the gradient magnitudes of the model updates (see Section \ref{sec:dp})\added{ and compare it to when all clients use DP-SGD as their local training optimizer}. \replaced{As an illustrative example, we first evaluate the simple DP protocol}{We evaluate this technique} on client 9, which is the high-risk client, using model updates that are sent at round 90. As shown in Fig.~\ref{fig:vary_dp}, the image reconstruction quality, as measured by the SSIM between original and recovered images, expectedly degrades as more noise is added to model updates from the client.

\begin{figure}[htbp]
    \centering
    \includegraphics[width=0.95\columnwidth]{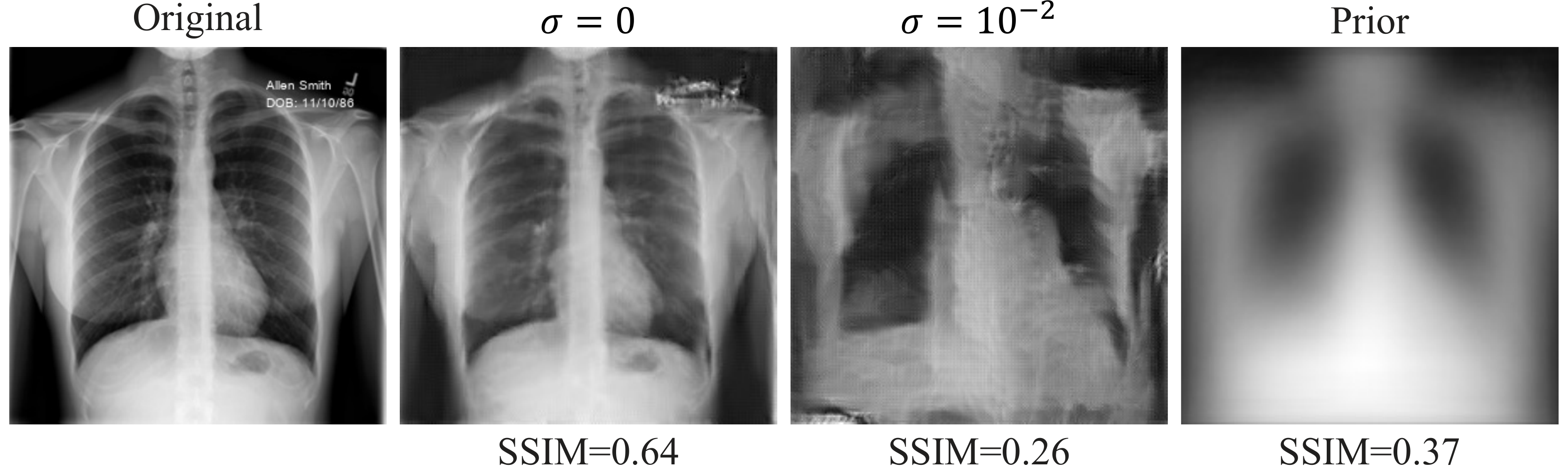}
    \caption{\textbf{Structural similarity (SSIM).} SSIM can be used to measure both the similarity between an original image and reconstruction or the similarity between the prior image used in the attack and the original image. An SSIM value below the one compared with the prior could indicate that no useful information was leaked or is recoverable by the attack. Note: the patient's name and date of birth in the original image are randomly generated. \label{fig:ssim_comparison}}
\end{figure}

The SSIM value between the prior which is used to initialize the gradient inversion attack (see \ref{sec:practical_attack}) and the original image is 0.37 (see Fig. \ref{fig:ssim_comparison}). This can be considered as a lower bound of this metric during the attack. In addition, as the noise level increases, reconstructions from model updates with added DP become closer to the prior image and less similar to the original training image. This indicates that a privacy setting ($\sigma$) allowing only a reconstruction with a SSIM value equal to or less than that between the original image and the prior could be considered ``safe''. For instance, as shown in Fig. \ref{fig:ssim_comparison}, this behavior can be observed for the high-risk client 9 at FL round 90 and noise added with $\sigma = 10^{-2}$.

\subsection{Quantification of Data Leakage}
\label{sec:quantify_leakage}

We further visualize the degree of data leakage across different clients in FL with respect to different levels of privacy preservation through added Gaussian noise\added{, both in the simple DP protocol and under DP-SGD}. Comparing data leakage across clients with different images is not straightforward. This is due to the variation in the reference images used to compute the SSIM, which renders the absolute values of the similarity metrics between the reconstruction and the training images incomparable. Therefore, we utilize a relative change with respect to a reference point to compute a more comparable quantitative metric across different clients. We define a new metric denoted as ``Relative Data Leakage Value'' (RDLV) to measure the relative change of the similarity metric (SSIM) for reconstructions from different clients' model updates. Given a training image $T_i$, the prior used in the gradient inversion attack $P$, and the corresponding reconstruction $I_i$, RDLV is defined as:   

\begin{equation}
    \mathrm{RDLV} = \frac{\mathrm{SSIM}(T_i,I_i)-\mathrm{SSIM}(T_i,P)}{\mathrm{SSIM}(T_i,P)}
\end{equation}

\begin{figure}[htbp]
\begin{subfigure}{1.0\columnwidth}
    \centering
    \includegraphics[width=0.95\columnwidth]{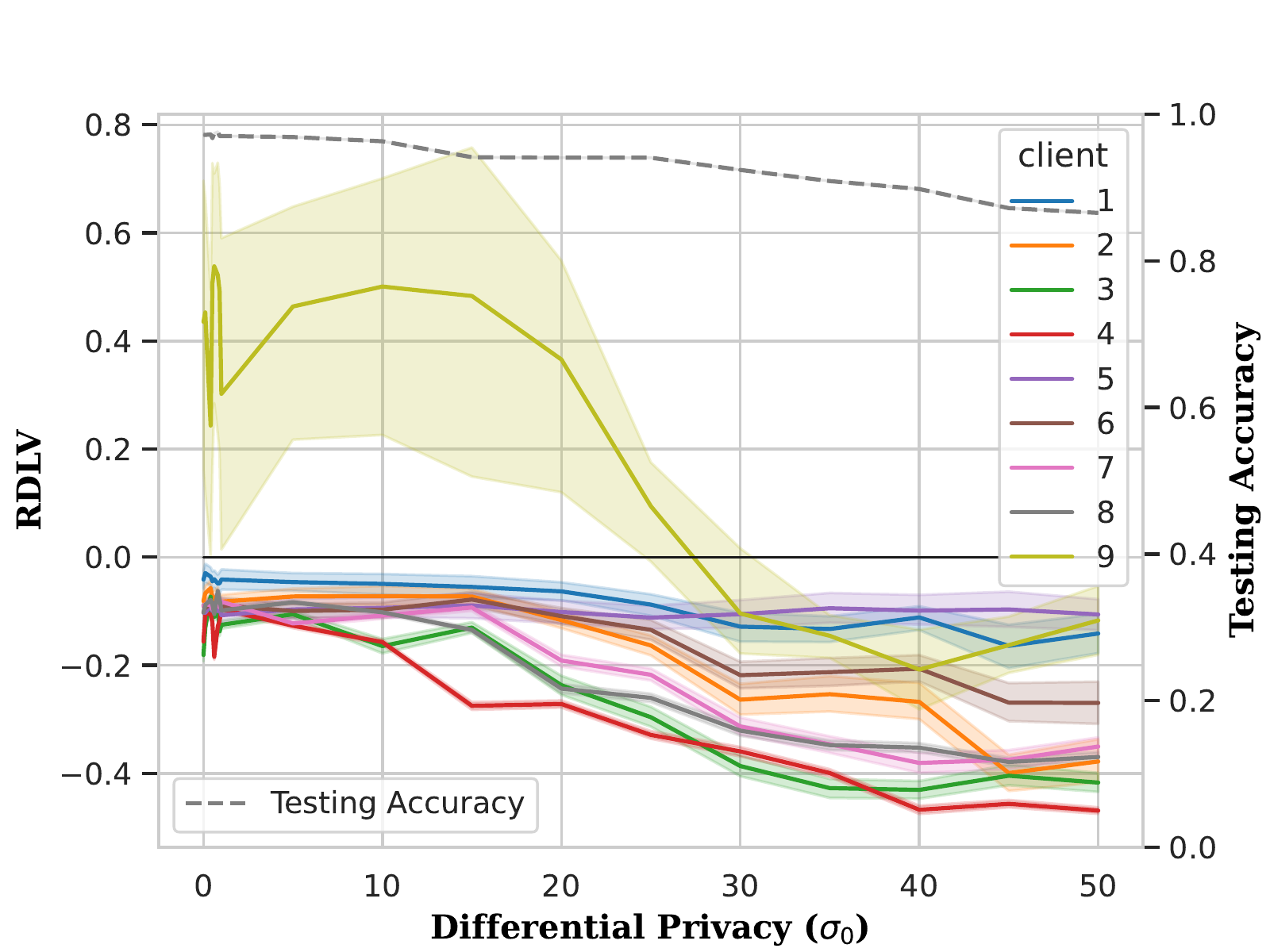}
    \caption{Chest X-ray images}
\end{subfigure}
\begin{subfigure}{1.0\columnwidth}
    \centering
    \includegraphics[width=0.95\columnwidth]{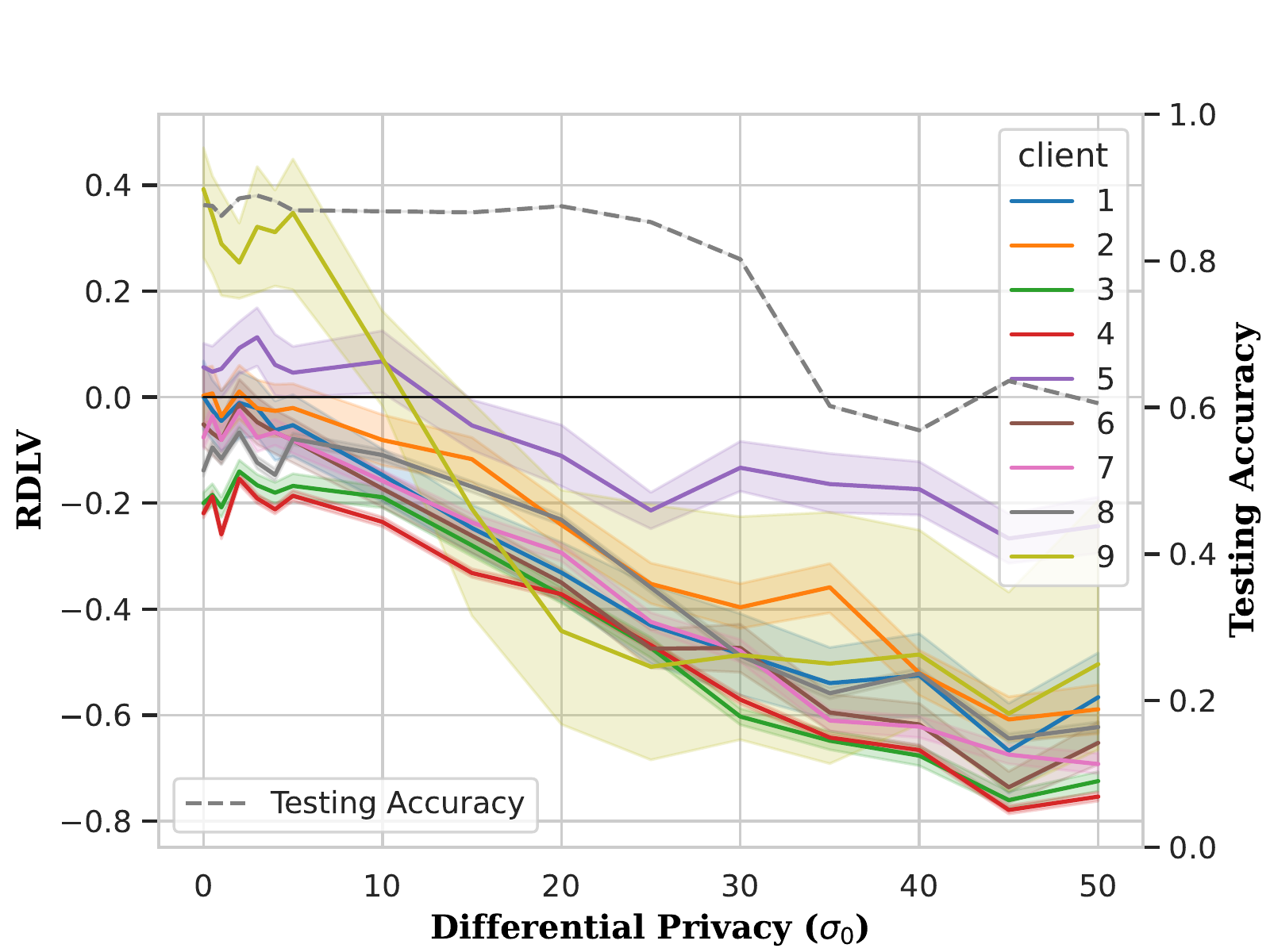}
    \caption{Fundus images}
\end{subfigure}
\caption{\textbf{Quantification of data leakage.} ``Relative Data Leakage Value'' (RDLV) for each client compared to the prior used in the gradient inversion attacks for different DP settings ($\sigma_0$) in the chest X-ray (a) and fundus image application (b). Higher $\sigma_0$ values add more noise to obfuscate the model updates sent by the clients. The testing accuracy of the best global model is shown for comparison. 95\% confidence interval bands are shown for each client. \label{fig:rdlv}}
\end{figure}

Fig.~\ref{fig:rdlv} illustrates the RDLV based on different differential privacy settings for each client. Each line shows the average central tendency and a 95\% confidence interval (bands shown in the plot) estimated using bootstrapping (with 1000 trials) on reconstructions that achieve the highest SSIM with any of the original training images during different rounds of FL. As observed, only the high-risk client 9, which sends model updates from training one iteration on one image, leaks the training data up to a certain amount of noise which corresponds to a positive value of RDLV (up to around $\sigma_0 \geq 25$ for most FL rounds in this case). A positive relative change indicates that the reconstructions are more similar to the original images than the prior used in the gradient inversion attack. In addition, clients with more than one training image do not leak data in this regime within the confidence interval as their reconstructions are less similar than the prior used in the attack. As expected, the global model accuracy drops as more noise is added to the model updates from clients during FL.

\begin{figure}[htbp]
  \centering
  \includegraphics[width=0.95\columnwidth]{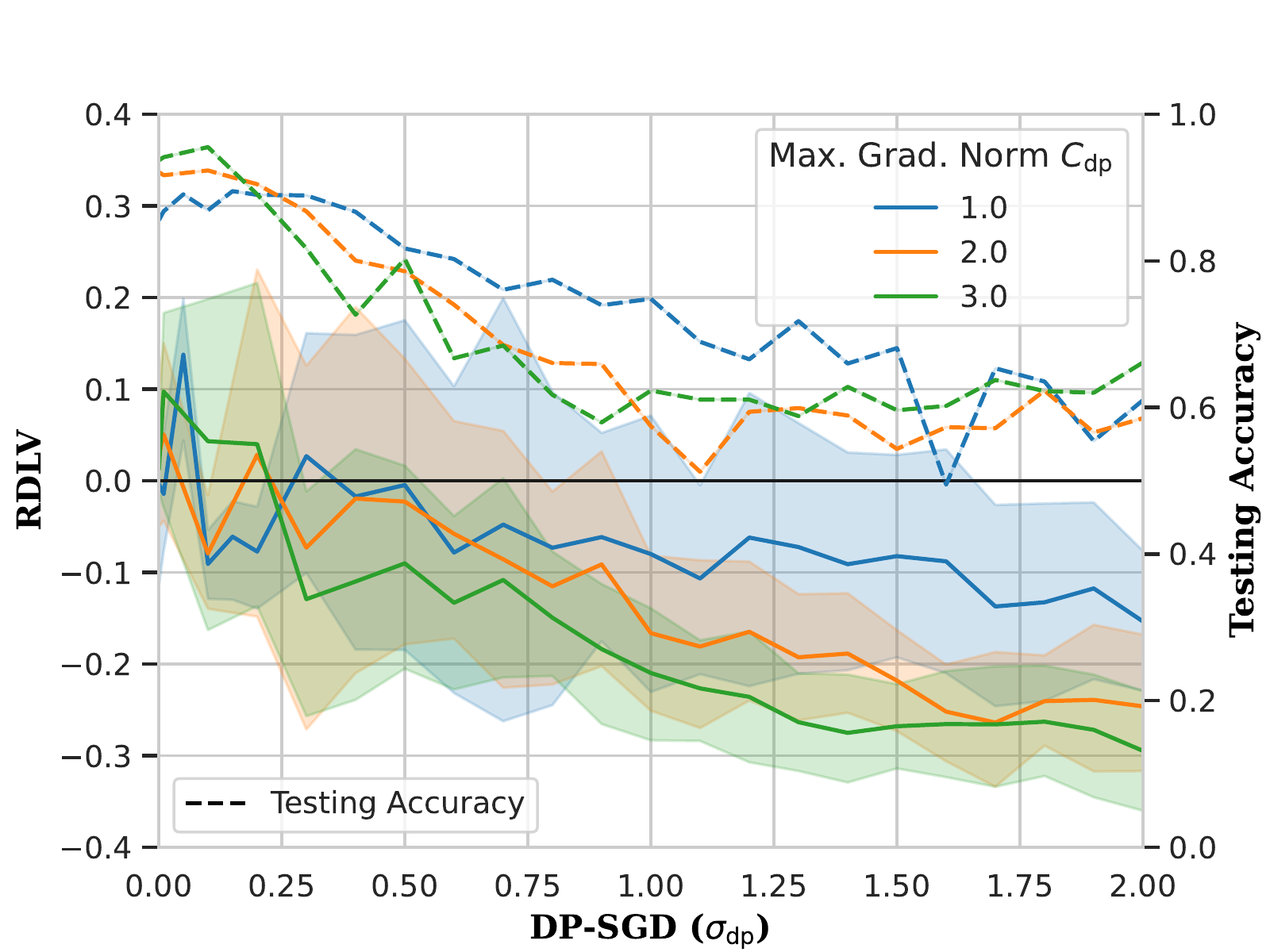}
\caption{\added{\textbf{Quantification of data leakage under DP-SGD (Chest X-ray images).} ``Relative Data Leakage Value'' (RDLV) for the ``high-risk'' client 9 under different privacy settings of DP-SGD.} \label{fig:dp-sgd}}
\end{figure}

%
%
\added{Fig.~\ref{fig:dp-sgd} shows the RDLV of the ``high-risk'' client 9 under different privacy settings of DP-SGD using different $l_{2}$ norm gradient clipping values $C_\mathrm{dp}$ and noise multiplier values $\sigma_\mathrm{dp}$. Otherwise, the experimental settings are identical to the above experiments. One can observe that DP-SGD is more detrimental to the global model performance compared to the simple Gaussian noise protocol presented in Fig.~\ref{fig:rdlv}. However, DP-SGD is also effectively reducing the data leakage (achieving negative RDLV) for the ``high-risk'' client. Additional tuning of the privacy parameters specific to each client might be needed to optimize the global model performance under DP-SGD.}

\subsection{Deep Embedding-Based Visualizations}
\label{sec:tsne_embeddings}

\begin{figure*}[htbp]
\centering
\begin{tabular}{c|c}
\begin{subfigure}[b]{.295\textwidth}
    \includegraphics[width=\textwidth]{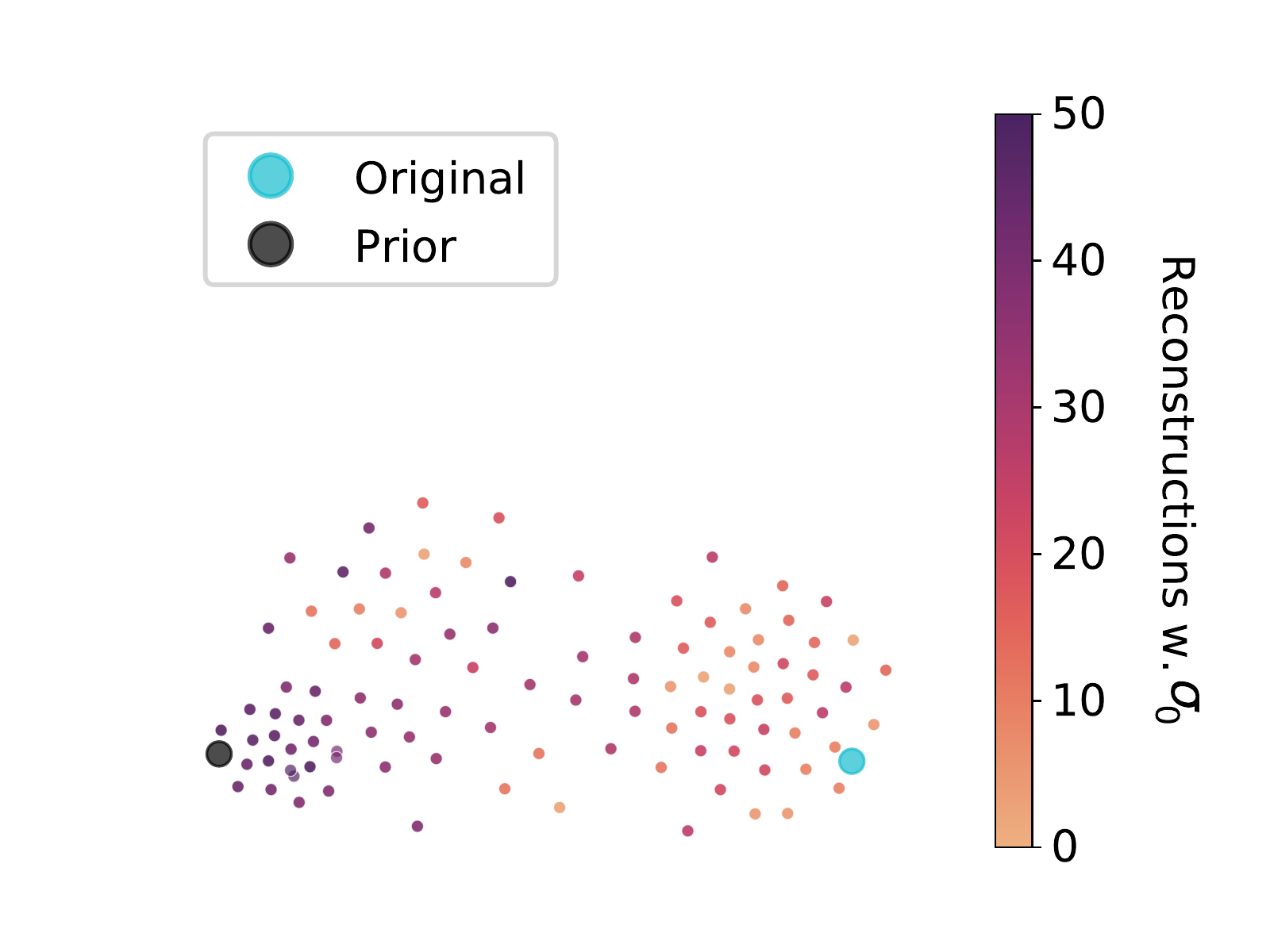}
    \caption{High-risk client 9 \label{fig:tsne_a}}    
\end{subfigure}
&
\begin{subfigure}[b]{.695\textwidth}
    \includegraphics[width=\textwidth]{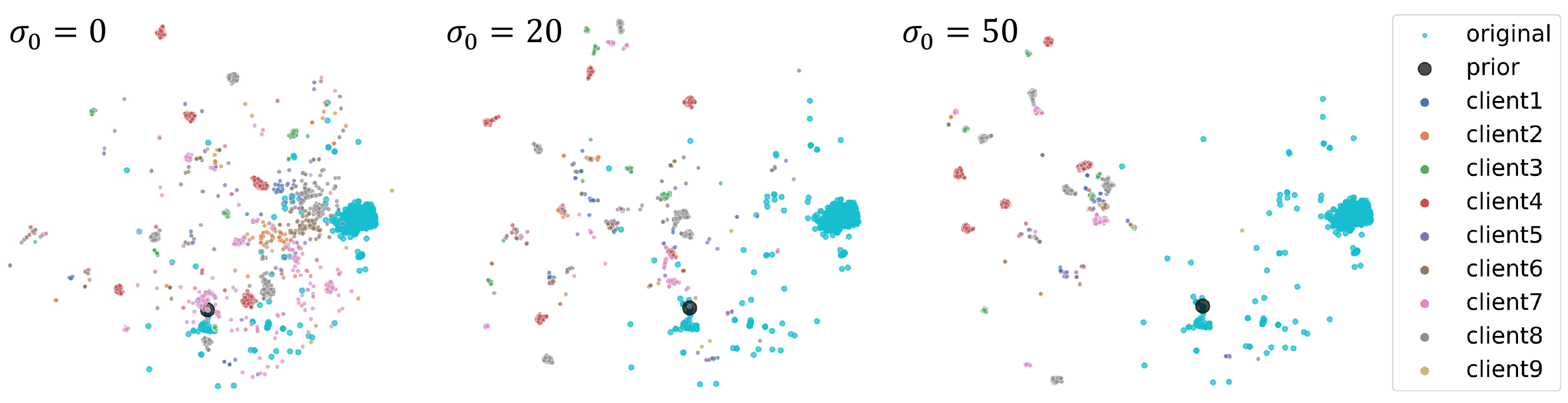}
    \caption{All clients \label{fig:tsne_b}}
\end{subfigure}
\end{tabular}
\caption{\textbf{Visualizing data leakage.} Two-dimensional t-SNE~\cite{van2008visualizing} embeddings of deep features extracted from original training images and reconstructions at different DP levels as indicated by $\sigma_0$.
The colorbar shows the $\sigma_0$ value for the corresponding reconstructions.
\label{fig:tsne}}
\end{figure*}

Alternative ways to measure the similarity between reconstructed images and training images are based on deep feature embeddings. We use an off-the-shelf ImageNet-pretrained ResNet-18~\cite{he2016deep} to extract 512-dimensional feature embeddings from training and reconstructed images. In Fig. \ref{fig:tsne}, we visualize the deep feature embeddings of training images and reconstructions using t-SNE~\cite{van2008visualizing}. As shown, with higher values of Gaussian noise for privacy preservation, the clusters of original images and reconstructions further drift apart as the similarity between the original training images and any of the reconstructions decrease significantly.

\subsection{Membership Inference Attack Data Leakage}
\label{sec:inference_attacks}

\begin{figure*}[htbp]
    \centering
    \includegraphics[width=1.8\columnwidth]{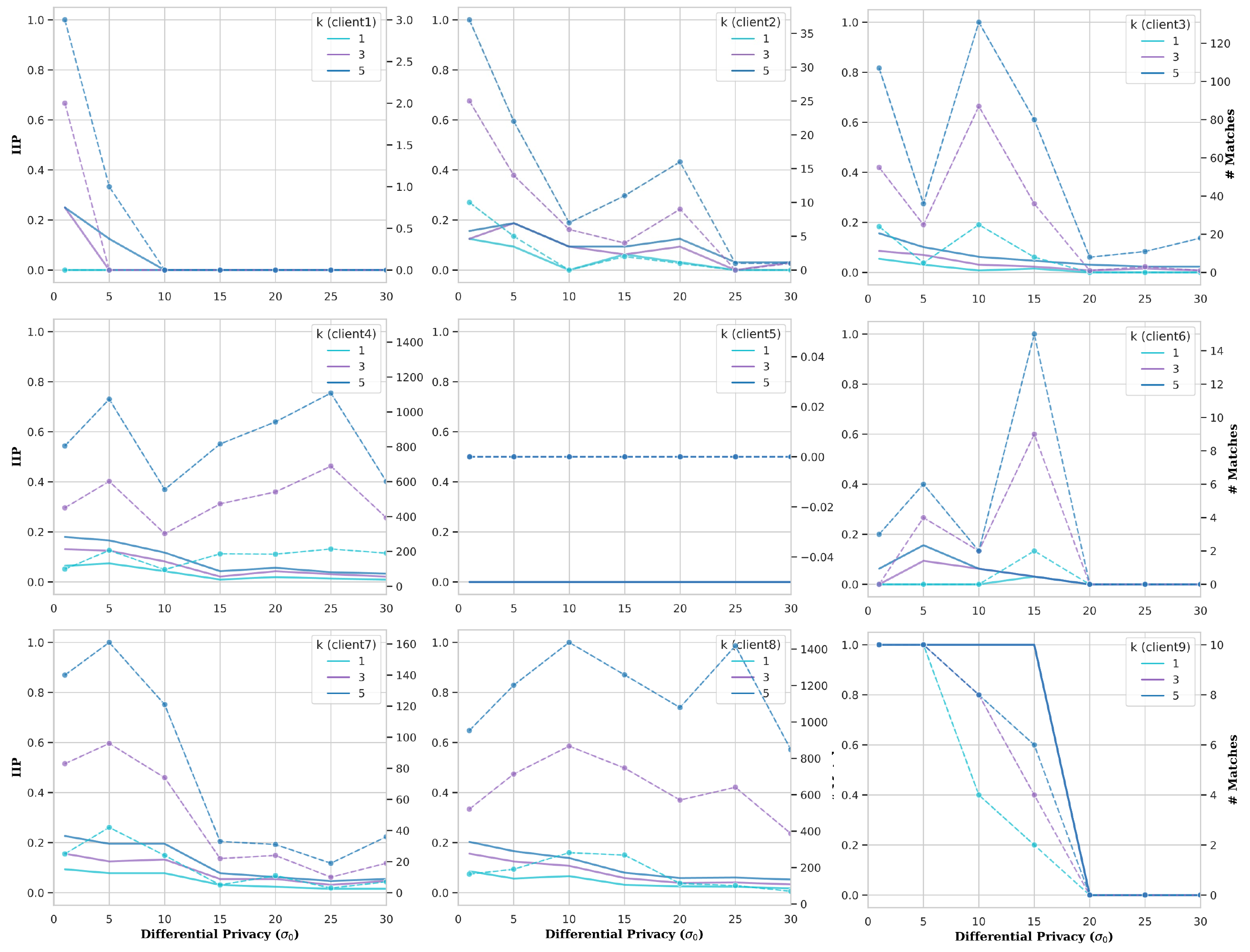}
    \caption{\textbf{k-Image Identifiability Precision (k-IIP).} k-IIP (solid lines) measures how many reconstructions could be used to identify the exact members of the training images that are used during FL for a specific client using different differential privacy settings ($\sigma_0$). We also show the number of matches (dotted lines) of feature embeddings between the reconstructions and original images used to identify the exact members for all clients (Only unique matches are used to compute the k-IIP value). Only the cosine similarity of matched exact members is shown (IIP $>$ 0). Both IIP and cosine similarities reduce with higher Gaussian noise levels and are typically lower for clients with larger amounts of training images. \label{fig:iip_attack}}
\end{figure*}

We furthermore adjusted the Image Identifiability Precision (IIP) proposed in Yin et al.~\cite{yin2021see} to the FL scenario. Our specific approach is outlined in Section \ref{sec:iip}. IIP measures the amount of identifiable information leaked in the gradient inversion attack by counting the closest exact matches between the reconstructions from the membership attack and the original training images for each client. As shown in Fig. \ref{fig:iip_attack}, the high-risk client 9 (updates sent based on 1 image and 1 iteration) attains the highest IIP score of 1.0 up to a Gaussian noise level $\sigma_0$ value of $\sim$20. This indicates that the attacker is able to determine the exact image that participated in FL training for that specific client. Furthermore, all other clients already achieve significantly lower values of the IIP score (e.g., $<$0.1) with much lower Gaussian noise levels $\sigma_0$. We also show the cosine similarity (dotted lines) between the reconstructions and original images identifying exact members of the training set at each client based on ImageNet-pretrained ResNet-18~\cite{he2016deep} feature embeddings. As expected, cosine similarity values also decrease with higher Gaussian noise levels. 

\section{Discussion}
\label{sec:discussion}


In this work, we have introduced a new way to measure the data leakage in horizontal FL for chest X-ray image classification and proposed several ways to quantify and visualize it. Overall, we conclude that FL security is multifaceted. Clients which send updates based on a small number of training images are at a higher risk of leaking data than the clients sending updates based on larger training sets as well as higher numbers of local training iterations. Furthermore, larger batch sizes make reconstruction via gradient inversion attacks more difficult, a finding that is in line with previous literature~\cite{geiping2020inverting,kaissis2021end,yin2021see}. 
The following summarizes our overall conclusions based on the findings shown in this study.

\paragraph{What is insecure?}                        
\begin{itemize}
    \item ↓ small training sets                    
    \item ↓ updates from small number of iterations
    \item ↓ small batch sizes                     
\end{itemize}

\paragraph{What is ``more'' secure?}           
\begin{itemize}
    \item ↑ large training sets                       
    \item ↑ updates from a large number of iterations over different
    \item ↑ large batch sizes                                        
\end{itemize}

While sending model updates using a single training image is at a high risk of data leakage, clients that utilize more realistic settings, such as larger number of training images and large number of local iterations as common in real-world FL scenarios, are not at a high risk of leaking identifiable information. Specifically, gradient inversion attacks for clients that iterate over several images before sending model updates are harder due to accumulated errors in the estimation of BN statistics. In this scenario, the attack is executed using only an approximation of the BN statistics (see Fig. \ref{fig:reconstructions}, Fig.~\ref{fig:vary_batchsize}, and Section \ref{sec:practical_attack}). 

In addition, reconstructions from model updates that are intercepted in later stages of FL training exhibit higher image quality in terms of image fidelity and capturing the fine-grained details of anatomical structure. This improved image generation capability could be attributed to the improved feature learning capability of the global model in later rounds of training. However, these images still do not bear any identifiable similarities to the individual training images. As shown in Fig. \ref{fig:reconstructions}, the reconstructed images start to resemble real chest X-ray images, but none of them could be found in the original training set. As a result, such reconstructions do not constitute a significant security breach.
In contrast, models trained from scratch are less likely to leak more informative gradients. However, pre-training is common practice in modern deep learning, and not using well-pretrained models often results in a significant drop in performance.

Additionally, there are several potential mitigations against the server-side gradient inversion attacks that could be easily incorporated into modern FL systems:

\begin{enumerate}
\item If the attacker does not have access to the current state of the global model with respect to which the client’s updates are computed, then the attack becomes unsuccessful (see Fig. \ref{fig:attack_success_c}). In our experience, removing access to the state of only the most important layers can render the attack ineffective. Specifically, this could be achieved efficiently by utilizing cryptographic techniques such as homomorphic encryption to allow secure aggregation of model updates~\cite{zhang2020batchcrypt}. Encryption would ensure that the server never holds the complete unencrypted model to avoid a potential gradient inversion attack while adding negligible computing requirements.

\item If the attacker does not have access to the model architecture, attempting to match the gradients will become very challenging. The clients could be sending only numerical arrays of model updates to be aggregated on the server without the server ever holding the information about the current model architecture or training procedure. While theoretically, the attacker could infer common architectures from the size of the parameter arrays, in practice, it would become extremely difficult, if not impossible, to deduce the exact architecture, especially for customized architectures often used in healthcare applications.

\item For models that are trained with BN, not sending the BN statistics to the server causes the gradient inversion attack to fail (see Fig. \ref{fig:attack_success}. Techniques such as FedBN~\cite{li2021fedbn,andreux2020siloed} could reduce the risk of server-side attacks by keeping personalized BN statistics specific to each client. One negative aspect of this approach is the lack of a common global model after FL training since only an ensemble of personalized client models would be available. 
\end{enumerate}

In addition, DP techniques typically reduce the risks of data leakage in many scenarios. In this work, we explored a simple variation of the Gaussian DP mechanism to add calibrated noise to the model updates before sending them to the server \added{and compared it to DP-SGD}. Even \replaced{the simple DP protocol}{this simple approach} showed to be effective in reducing the risks of gradient inversion attacks. However, a noticeable degradation in model performance was observed with higher Gaussian noise levels. \added{Model degradation was even more pronounced with higher levels of privacy using DP-SGD}. As DP has shown to be an effective approach in reducing data leakage for gradient inversion attacks, future research efforts are needed to explore the optimal tradeoff between various DP mechanisms and model performance.   

Other techniques such as secure authentication, trusted execution environment, and secure communication using SSL encryption, and cryptographic methodologies such as homomorphic encryption~\cite{zhang2020batchcrypt} or secure multi-party computation~\cite{kaissis2020secure} to completely encrypt the exchanged updates could be used to ensure a secure FL. As previously discussed, targeted encryption of the most vulnerable parts of the model might be enough to ensure the security of training while only introducing a minimal computational overhead.

We also identified some limitations of our study. For instance, we only studied modern CNN architectures such as ResNet-18 that are commonly used in image classification tasks. We did not consider other network architectures such as those used for image segmentation or natural language processing in our analysis. Furthermore, we limited this study to clients using SGD as the training optimizer. More complicated optimization schemes such as FedOpt~\cite{reddi2020adaptive} might require further considerations. Moreover, using the previously discussed RDLV regime (positive relative change, see Fig. \ref{fig:rdlv}), some reconstructed images did not fall within the 95\% confidence interval, and as a result, could indicate ``data leakage''. However, qualitative comparisons did not reveal any data leakage or individually identifiable information in those reconstructions. Specifically, these reconstructions have a low IIP score and do not bear close similarities to the original training images. However, they have some qualities that cause a relatively high SSIM metric. With such cases, similarity metrics that are computed based on deep feature embeddings might provide more plausible values for comparison. 
\added{Alternative attacks that include the utilization of generative methods, e.g., GAN~\cite{hitaj2017deep,wang2019beyond}, have not been investigated in this study.}

Future work might identify the vulnerable populations such as outliers or individual images that might cause higher losses in later stages of FL training causing higher SSIM or IIP values. 
\deleted{Alternative attacks could include the utilization of generative methods, e.g. GAN, and need to be investigated.}
Furthermore, other applications such as 2D/3D segmentation and 3D classification along with novel model architectures should be investigated. 
Other attack scenarios like model inversion after FL training, as well as backdoor and membership attacks, should be further studied~\cite{usynin2021adversarial}.

In conclusion, we presented a new way to measure and visualize data leakage in real-world FL scenarios. Without resorting to contrived settings for enabling the gradient inversion attack~\cite{geiping2020inverting,kaissis2021end}, we studied a practical scenario in which model updates from clients were sent by accounting for updated BN statistics (i.e., training mode). We conclude that previous efforts likely exaggerated the risks of gradient inversion attacks in FL, and that our work builds a foundation towards a better understanding of the real risks of such attacks in real-world FL applications. In most cases, it becomes very difficult for an attacker to recover any training images, as common FL scenarios involve model updates from sufficiently large numbers of batch sizes, training images, and local iterations that detrimentally impact the success of current gradient inversion attacks. Furthermore, we listed several effective protection methods that could be easily incorporated into any FL system to render existing attacks futile.
We believe that our work is a step towards establishing reproducible methods of measuring data leakage in FL and therefore builds a framework for finding an optimal tradeoff between privacy and accuracy based on quantifiable metrics. 


\bibliographystyle{IEEEtran}
\bibliography{refs}

\begin{thebibliography}{10}
\providecommand{\url}[1]{#1}
\csname url@samestyle\endcsname
\providecommand{\newblock}{\relax}
\providecommand{\bibinfo}[2]{#2}
\providecommand{\BIBentrySTDinterwordspacing}{\spaceskip=0pt\relax}
\providecommand{\BIBentryALTinterwordstretchfactor}{4}
\providecommand{\BIBentryALTinterwordspacing}{\spaceskip=\fontdimen2\font plus
\BIBentryALTinterwordstretchfactor\fontdimen3\font minus
  \fontdimen4\font\relax}
\providecommand{\BIBforeignlanguage}[2]{{%
\expandafter\ifx\csname l@#1\endcsname\relax
\typeout{** WARNING: IEEEtran.bst: No hyphenation pattern has been}%
\typeout{** loaded for the language `#1'. Using the pattern for}%
\typeout{** the default language instead.}%
\else
\language=\csname l@#1\endcsname
\fi
#2}}
\providecommand{\BIBdecl}{\relax}
\BIBdecl

\bibitem{rieke2020future}
N.~Rieke, J.~Hancox, W.~Li, F.~Milletari, H.~R. Roth, S.~Albarqouni, S.~Bakas,
  M.~N. Galtier, B.~A. Landman, K.~Maier-Hein \emph{et~al.}, ``The future of
  digital health with federated learning,'' \emph{NPJ digital medicine},
  vol.~3, no.~1, pp. 1--7, 2020.

\bibitem{mcmahan2017communication}
B.~McMahan, E.~Moore, D.~Ramage, S.~Hampson, and B.~A. y~Arcas,
  ``Communication-efficient learning of deep networks from decentralized
  data,'' in \emph{Artificial intelligence and statistics}.\hskip 1em plus
  0.5em minus 0.4em\relax PMLR, 2017, pp. 1273--1282.

\bibitem{sheller2018multi}
M.~J. Sheller, G.~A. Reina, B.~Edwards, J.~Martin, and S.~Bakas,
  ``Multi-institutional deep learning modeling without sharing patient data: A
  feasibility study on brain tumor segmentation,'' in \emph{International
  MICCAI Brainlesion Workshop}.\hskip 1em plus 0.5em minus 0.4em\relax
  Springer, 2018, pp. 92--104.

\bibitem{roth2020federated}
H.~R. Roth, K.~Chang, P.~Singh, N.~Neumark, W.~Li, V.~Gupta, S.~Gupta, L.~Qu,
  A.~Ihsani, B.~C. Bizzo \emph{et~al.}, ``Federated learning for breast density
  classification: A real-world implementation,'' in \emph{Domain Adaptation and
  Representation Transfer, and Distributed and Collaborative Learning}.\hskip
  1em plus 0.5em minus 0.4em\relax Springer, 2020, pp. 181--191.

\bibitem{sheller2020federated}
M.~J. Sheller, B.~Edwards, G.~A. Reina, J.~Martin, S.~Pati, A.~Kotrotsou,
  M.~Milchenko, W.~Xu, D.~Marcus, R.~R. Colen \emph{et~al.}, ``Federated
  learning in medicine: facilitating multi-institutional collaborations without
  sharing patient data,'' \emph{Scientific reports}, vol.~10, no.~1, pp. 1--12,
  2020.

\bibitem{remedios2020federated}
S.~W. Remedios, J.~A. Butman, B.~A. Landman, and D.~L. Pham, ``Federated
  gradient averaging for multi-site training with momentum-based optimizers,''
  in \emph{Domain Adaptation and Representation Transfer, and Distributed and
  Collaborative Learning}.\hskip 1em plus 0.5em minus 0.4em\relax Springer,
  2020, pp. 170--180.

\bibitem{sarma2021federated}
K.~V. Sarma, S.~Harmon, T.~Sanford, H.~R. Roth, Z.~Xu, J.~Tetreault, D.~Xu,
  M.~G. Flores, A.~G. Raman, R.~Kulkarni \emph{et~al.}, ``Federated learning
  improves site performance in multicenter deep learning without data
  sharing,'' \emph{Journal of the American Medical Informatics Association},
  vol.~28, no.~6, pp. 1259--1264, 2021.

\bibitem{dayan2021federated}
I.~Dayan, H.~R. Roth, A.~Zhong, A.~Harouni, A.~Gentili, A.~Z. Abidin, A.~Liu,
  A.~B. Costa, B.~J. Wood, C.-S. Tsai \emph{et~al.}, ``Federated learning for
  predicting clinical outcomes in patients with covid-19,'' \emph{Nature
  medicine}, vol.~27, no.~10, pp. 1735--1743, 2021.

\bibitem{kairouz2019advances}
P.~Kairouz, H.~B. McMahan, B.~Avent, A.~Bellet, M.~Bennis, A.~N. Bhagoji,
  K.~Bonawitz, Z.~Charles, G.~Cormode, R.~Cummings \emph{et~al.}, ``Advances
  and open problems in federated learning,'' \emph{arXiv preprint
  arXiv:1912.04977}, 2019.

\bibitem{li2020federated}
T.~Li, A.~K. Sahu, M.~Zaheer, M.~Sanjabi, A.~Talwalkar, and V.~Smith,
  ``Federated optimization in heterogeneous networks,'' \emph{Proceedings of
  Machine Learning and Systems}, vol.~2, pp. 429--450, 2020.

\bibitem{karimireddy2020scaffold}
S.~P. Karimireddy, S.~Kale, M.~Mohri, S.~Reddi, S.~Stich, and A.~T. Suresh,
  ``Scaffold: Stochastic controlled averaging for federated learning,'' in
  \emph{International Conference on Machine Learning}.\hskip 1em plus 0.5em
  minus 0.4em\relax PMLR, 2020, pp. 5132--5143.

\bibitem{zhou2021towards}
Z.~Zhou, L.~Chu, C.~Liu, L.~Wang, J.~Pei, and Y.~Zhang, ``Towards fair
  federated learning,'' in \emph{Proceedings of the 27th ACM SIGKDD Conference
  on Knowledge Discovery \& Data Mining}, 2021, pp. 4100--4101.

\bibitem{rothchild2020fetchsgd}
D.~Rothchild, A.~Panda, E.~Ullah, N.~Ivkin, I.~Stoica, V.~Braverman,
  J.~Gonzalez, and R.~Arora, ``Fetchsgd: Communication-efficient federated
  learning with sketching,'' in \emph{International Conference on Machine
  Learning}.\hskip 1em plus 0.5em minus 0.4em\relax PMLR, 2020, pp. 8253--8265.

\bibitem{geiping2020inverting}
J.~Geiping, H.~Bauermeister, H.~Dr{\"o}ge, and M.~Moeller, ``Inverting
  gradients-how easy is it to break privacy in federated learning?''
  \emph{Advances in Neural Information Processing Systems}, vol.~33, pp.
  16\,937--16\,947, 2020.

\bibitem{yin2021see}
H.~Yin, A.~Mallya, A.~Vahdat, J.~M. Alvarez, J.~Kautz, and P.~Molchanov, ``See
  through gradients: Image batch recovery via gradinversion,'' in
  \emph{Proceedings of the IEEE/CVF Conference on Computer Vision and Pattern
  Recognition}, 2021, pp. 16\,337--16\,346.

\bibitem{kaissis2021end}
G.~Kaissis, A.~Ziller, J.~Passerat-Palmbach, T.~Ryffel, D.~Usynin, A.~Trask,
  I.~Lima, J.~Mancuso, F.~Jungmann, M.-M. Steinborn \emph{et~al.}, ``End-to-end
  privacy preserving deep learning on multi-institutional medical imaging,''
  \emph{Nature Machine Intelligence}, vol.~3, no.~6, pp. 473--484, 2021.

\bibitem{bonawitz2017practical}
K.~Bonawitz, V.~Ivanov, B.~Kreuter, A.~Marcedone, H.~B. McMahan, S.~Patel,
  D.~Ramage, A.~Segal, and K.~Seth, ``Practical secure aggregation for
  privacy-preserving machine learning,'' in \emph{proceedings of the 2017 ACM
  SIGSAC Conference on Computer and Communications Security}, 2017, pp.
  1175--1191.

\bibitem{yang2019federated}
Q.~Yang, Y.~Liu, T.~Chen, and Y.~Tong, ``Federated machine learning: Concept
  and applications,'' \emph{ACM Transactions on Intelligent Systems and
  Technology (TIST)}, vol.~10, no.~2, pp. 1--19, 2019.

\bibitem{ioffe2015batch}
S.~Ioffe and C.~Szegedy, ``Batch normalization: Accelerating deep network
  training by reducing internal covariate shift,'' in \emph{International
  conference on machine learning}.\hskip 1em plus 0.5em minus 0.4em\relax PMLR,
  2015, pp. 448--456.

\bibitem{zhu2019deep}
L.~Zhu, Z.~Liu, and S.~Han, ``Deep leakage from gradients,'' \emph{Advances in
  Neural Information Processing Systems}, vol.~32, 2019.

\bibitem{he2016deep}
K.~He, X.~Zhang, S.~Ren, and J.~Sun, ``Deep residual learning for image
  recognition,'' in \emph{Proceedings of the IEEE conference on computer vision
  and pattern recognition}, 2016, pp. 770--778.

\bibitem{huang2017densely}
G.~Huang, Z.~Liu, L.~Van Der~Maaten, and K.~Q. Weinberger, ``Densely connected
  convolutional networks,'' in \emph{Proceedings of the IEEE conference on
  computer vision and pattern recognition}, 2017, pp. 4700--4708.

\bibitem{fredrikson2015model}
M.~Fredrikson, S.~Jha, and T.~Ristenpart, ``Model inversion attacks that
  exploit confidence information and basic countermeasures,'' in
  \emph{Proceedings of the 22nd ACM SIGSAC conference on computer and
  communications security}, 2015, pp. 1322--1333.

\bibitem{yin2020dreaming}
H.~Yin, P.~Molchanov, J.~M. Alvarez, Z.~Li, A.~Mallya, D.~Hoiem, N.~K. Jha, and
  J.~Kautz, ``Dreaming to distill: Data-free knowledge transfer via
  deepinversion,'' in \emph{Proceedings of the IEEE/CVF Conference on Computer
  Vision and Pattern Recognition}, 2020, pp. 8715--8724.

\bibitem{dwork2006calibrating}
C.~Dwork, F.~McSherry, K.~Nissim, and A.~Smith, ``Calibrating noise to
  sensitivity in private data analysis,'' in \emph{Theory of cryptography
  conference}.\hskip 1em plus 0.5em minus 0.4em\relax Springer, 2006, pp.
  265--284.

\bibitem{yang2020local}
M.~Yang, L.~Lyu, J.~Zhao, T.~Zhu, and K.-Y. Lam, ``Local differential privacy
  and its applications: A comprehensive survey,'' \emph{arXiv preprint
  arXiv:2008.03686}, 2020.

\bibitem{lyu2016understanding}
M.~Lyu, D.~Su, and N.~Li, ``Understanding the sparse vector technique for
  differential privacy,'' \emph{arXiv preprint arXiv:1603.01699}, 2016.

\bibitem{abadi2016deep}
M.~Abadi, A.~Chu, I.~Goodfellow, H.~B. McMahan, I.~Mironov, K.~Talwar, and
  L.~Zhang, ``Deep learning with differential privacy,'' in \emph{Proceedings
  of the 2016 ACM SIGSAC conference on computer and communications security},
  2016, pp. 308--318.

\bibitem{wu2019p3sgd}
B.~Wu, S.~Zhao, G.~Sun, X.~Zhang, Z.~Su, C.~Zeng, and Z.~Liu, ``P3sgd: Patient
  privacy preserving sgd for regularizing deep cnns in pathological image
  classification,'' in \emph{Proceedings of the IEEE/CVF Conference on Computer
  Vision and Pattern Recognition}, 2019, pp. 2099--2108.

\bibitem{ziller2021medical}
A.~Ziller, D.~Usynin, R.~Braren, M.~Makowski, D.~Rueckert, and G.~Kaissis,
  ``Medical imaging deep learning with differential privacy,'' \emph{Scientific
  Reports}, vol.~11, no.~1, pp. 1--8, 2021.

\bibitem{li2019privacy}
W.~Li, F.~Milletar{\`\i}, D.~Xu, N.~Rieke, J.~Hancox, W.~Zhu, M.~Baust,
  Y.~Cheng, S.~Ourselin, M.~J. Cardoso \emph{et~al.}, ``Privacy-preserving
  federated brain tumour segmentation,'' in \emph{International workshop on
  machine learning in medical imaging}.\hskip 1em plus 0.5em minus 0.4em\relax
  Springer, 2019, pp. 133--141.

\bibitem{liang2020differentially}
Z.~Liang, B.~Wang, Q.~Gu, S.~Osher, and Y.~Yao, ``Differentially private
  federated learning with laplacian smoothing,'' \emph{arXiv preprint
  arXiv:2005.00218}, 2020.

\bibitem{rodriguez2020federated}
N.~Rodr{\'\i}guez-Barroso, G.~Stipcich, D.~Jim{\'e}nez-L{\'o}pez, J.~A.
  Ruiz-Mill{\'a}n, E.~Mart{\'\i}nez-C{\'a}mara, G.~Gonz{\'a}lez-Seco, M.~V.
  Luz{\'o}n, M.~A. Veganzones, and F.~Herrera, ``Federated learning and
  differential privacy: Software tools analysis, the sherpa. ai fl framework
  and methodological guidelines for preserving data privacy,''
  \emph{Information Fusion}, vol.~64, pp. 270--292, 2020.

\bibitem{adnan2022federated}
M.~Adnan, S.~Kalra, J.~C. Cresswell, G.~W. Taylor, and H.~R. Tizhoosh,
  ``Federated learning and differential privacy for medical image analysis,''
  \emph{Scientific Reports}, vol.~12, no.~1, pp. 1--10, 2022.

\bibitem{bagdasaryan2019differential}
E.~Bagdasaryan, O.~Poursaeed, and V.~Shmatikov, ``Differential privacy has
  disparate impact on model accuracy,'' \emph{Advances in Neural Information
  Processing Systems}, vol.~32, 2019.

\bibitem{song2020systematic}
L.~Song and P.~Mittal, ``Systematic evaluation of privacy risks of machine
  learning models,'' \emph{arXiv preprint arXiv:2003.10595}, 2020.

\bibitem{shokri2015privacy}
R.~Shokri and V.~Shmatikov, ``Privacy-preserving deep learning,'' in
  \emph{Proceedings of the 22nd ACM SIGSAC conference on computer and
  communications security}, 2015, pp. 1310--1321.

\bibitem{zhang2020batchcrypt}
C.~Zhang, S.~Li, J.~Xia, W.~Wang, F.~Yan, and Y.~Liu, ``Batchcrypt: Efficient
  homomorphic encryption for $\{$Cross-Silo$\}$ federated learning,'' in
  \emph{2020 USENIX Annual Technical Conference (USENIX ATC 20)}, 2020, pp.
  493--506.

\bibitem{naseri2020local}
M.~Naseri, J.~Hayes, and E.~De~Cristofaro, ``Local and central differential
  privacy for robustness and privacy in federated learning,'' \emph{arXiv
  preprint arXiv:2009.03561}, 2020.

\bibitem{mcmahan2018general}
H.~B. McMahan, G.~Andrew, U.~Erlingsson, S.~Chien, I.~Mironov, N.~Papernot, and
  P.~Kairouz, ``A general approach to adding differential privacy to iterative
  training procedures,'' \emph{arXiv preprint arXiv:1812.06210}, 2018.

\bibitem{sun2021soteria}
J.~Sun, A.~Li, B.~Wang, H.~Yang, H.~Li, and Y.~Chen, ``Soteria: Provable
  defense against privacy leakage in federated learning from representation
  perspective,'' in \emph{Proceedings of the IEEE/CVF Conference on Computer
  Vision and Pattern Recognition}, 2021, pp. 9311--9319.

\bibitem{deng2009imagenet}
J.~Deng, W.~Dong, R.~Socher, L.-J. Li, K.~Li, and L.~Fei-Fei, ``Imagenet: A
  large-scale hierarchical image database,'' in \emph{2009 IEEE conference on
  computer vision and pattern recognition}.\hskip 1em plus 0.5em minus
  0.4em\relax Ieee, 2009, pp. 248--255.

\bibitem{chowdhury2020can}
M.~E. Chowdhury, T.~Rahman, A.~Khandakar, R.~Mazhar, M.~A. Kadir, Z.~B. Mahbub,
  K.~R. Islam, M.~S. Khan, A.~Iqbal, N.~Al~Emadi \emph{et~al.}, ``Can ai help
  in screening viral and covid-19 pneumonia?'' \emph{IEEE Access}, vol.~8, pp.
  132\,665--132\,676, 2020.

\bibitem{rahman2021exploring}
T.~Rahman, A.~Khandakar, Y.~Qiblawey, A.~Tahir, S.~Kiranyaz, S.~B.~A. Kashem,
  M.~T. Islam, S.~Al~Maadeed, S.~M. Zughaier, M.~S. Khan \emph{et~al.},
  ``Exploring the effect of image enhancement techniques on covid-19 detection
  using chest x-ray images,'' \emph{Computers in biology and medicine}, vol.
  132, p. 104319, 2021.

\bibitem{wang2017chestx}
X.~Wang, Y.~Peng, L.~Lu, Z.~Lu, M.~Bagheri, and R.~M. Summers, ``Chestx-ray8:
  Hospital-scale chest x-ray database and benchmarks on weakly-supervised
  classification and localization of common thorax diseases,'' in
  \emph{Proceedings of the IEEE conference on computer vision and pattern
  recognition}, 2017, pp. 2097--2106.

\bibitem{de_vente_coen_2021_5793241}
\BIBentryALTinterwordspacing
C.~de~Vente, K.~A. Vermeer, N.~Jaccard, B.~van Ginneken, H.~G. Lemij, and C.~I.
  S{\'a}nchez, ``Rotterdam eyepacs airogs train set,'' Dec. 2021, {The previous
  version was split into two records. This new version contains all data and
  the second record is deprecated.} [Online]. Available:
  \url{https://doi.org/10.5281/zenodo.5793241}
\BIBentrySTDinterwordspacing

\bibitem{Roth_NVIDIA_FLARE_Federated_2022}
H.~R. Roth, Y.~Cheng, Y.~Wen, I.~Yang, Z.~Xu, Y.-T. Hsieh, K.~Kersten,
  A.~Harouni, C.~Zhao, K.~Lu, Z.~Zhang, W.~Li, A.~Myronenko, D.~Yang, S.~Yang,
  N.~Rieke, A.~Quraini, C.~Chen, D.~Xu, N.~Ma, P.~Dogra, M.~Flores, and
  A.~Feng, ``{NVIDIA FLARE: Federated Learning from Simulation to
  Real-World},'' \emph{International Workshop on Federated Learning, NeurIPS
  2022, New Orleans, USA}, 10 2022.

\bibitem{wang2004image}
Z.~Wang, A.~C. Bovik, H.~R. Sheikh, and E.~P. Simoncelli, ``Image quality
  assessment: from error visibility to structural similarity,'' \emph{IEEE
  transactions on image processing}, vol.~13, no.~4, pp. 600--612, 2004.

\bibitem{van2008visualizing}
L.~Van~der Maaten and G.~Hinton, ``Visualizing data using t-sne.''
  \emph{Journal of machine learning research}, vol.~9, no.~11, 2008.

\bibitem{li2021fedbn}
X.~Li, M.~Jiang, X.~Zhang, M.~Kamp, and Q.~Dou, ``Fedbn: Federated learning on
  non-iid features via local batch normalization,'' \emph{arXiv preprint
  arXiv:2102.07623}, 2021.

\bibitem{andreux2020siloed}
M.~Andreux, J.~O.~d. Terrail, C.~Beguier, and E.~W. Tramel, ``Siloed federated
  learning for multi-centric histopathology datasets,'' in \emph{Domain
  Adaptation and Representation Transfer, and Distributed and Collaborative
  Learning}.\hskip 1em plus 0.5em minus 0.4em\relax Springer, 2020, pp.
  129--139.

\bibitem{kaissis2020secure}
G.~A. Kaissis, M.~R. Makowski, D.~R{\"u}ckert, and R.~F. Braren, ``Secure,
  privacy-preserving and federated machine learning in medical imaging,''
  \emph{Nature Machine Intelligence}, vol.~2, no.~6, pp. 305--311, 2020.

\bibitem{reddi2020adaptive}
S.~Reddi, Z.~Charles, M.~Zaheer, Z.~Garrett, K.~Rush, J.~Kone{\v{c}}n{\`y},
  S.~Kumar, and H.~B. McMahan, ``Adaptive federated optimization,'' \emph{arXiv
  preprint arXiv:2003.00295}, 2020.

\bibitem{hitaj2017deep}
B.~Hitaj, G.~Ateniese, and F.~Perez-Cruz, ``Deep models under the gan:
  information leakage from collaborative deep learning,'' in \emph{Proceedings
  of the 2017 ACM SIGSAC conference on computer and communications security},
  2017, pp. 603--618.

\bibitem{wang2019beyond}
Z.~Wang, M.~Song, Z.~Zhang, Y.~Song, Q.~Wang, and H.~Qi, ``Beyond inferring
  class representatives: User-level privacy leakage from federated learning,''
  in \emph{IEEE INFOCOM 2019-IEEE Conference on Computer Communications}.\hskip
  1em plus 0.5em minus 0.4em\relax IEEE, 2019, pp. 2512--2520.

\bibitem{usynin2021adversarial}
D.~Usynin, A.~Ziller, M.~Makowski, R.~Braren, D.~Rueckert, B.~Glocker,
  G.~Kaissis, and J.~Passerat-Palmbach, ``Adversarial interference and its
  mitigations in privacy-preserving collaborative machine learning,''
  \emph{Nature Machine Intelligence}, vol.~3, no.~9, pp. 749--758, 2021.

\end{thebibliography}

\end{document}